\documentclass[letterpaper, 10 pt, conference]{ieeeconf}  

\IEEEoverridecommandlockouts                              

\usepackage{amsmath} 
\usepackage{amssymb}  
\usepackage[nolist]{acronym}
\usepackage{mathtools}

\usepackage{graphicx}
\usepackage{setspace}
\usepackage{epstopdf}
\usepackage{wrapfig}
\usepackage{subfig}
\usepackage{amssymb,amsmath}
\usepackage{color}
\usepackage[hyphens]{url}
\usepackage{amsmath}
\usepackage{algorithm}
\usepackage[noend]{algpseudocode}
\usepackage[english]{babel}
\usepackage[mathletters]{ucs}
\usepackage[utf8]{inputenc}
\usepackage[export]{adjustbox}
\usepackage{algorithm}
\usepackage{algpseudocode}

\usepackage[noadjust]{cite}

\algnewcommand\algnot{\textbf{not}}
\algnewcommand\algand{\textbf{and}}
\algnewcommand\algor{\textbf{or}}

\makeatletter
\newlength{\trianglerightwidth}
\algnewcommand{\LineComment}[1]{\Statex \hskip\ALG@thistlm $\triangleright$ #1}
\algnewcommand{\LeftComment}[1]{\Statex \(\triangleright\) #1}
\algnewcommand{\LineCommentCont}[1]{\State %
  \parbox[t]{\dimexpr\linewidth-\ALG@thistlm}{\hangindent=\trianglerightwidth \hangafter=1 \strut$\triangleright$ #1\strut}}
\makeatother

\usepackage{standalone}
\usepackage{latexsym}
\usepackage{indentfirst}
\usepackage{fancyhdr}
\usepackage{longtable}
\usepackage{pifont}
\usepackage{makeidx}
\usepackage{multirow}
\usepackage{dcolumn} 
\usepackage{listings}
\usepackage{relsize}
\usepackage{pdfpages}
\usepackage{tabularx,booktabs}
\newcolumntype{Y}{>{\arraybackslash}X}
\usepackage{booktabs}  
\usepackage[binary-units=true]{siunitx}
\usepackage{float}
\usepackage{textcomp}
\usepackage{bm}  
\usepackage{interval}
\usepackage{dblfloatfix}    

\usepackage{makecell}
\usepackage{lipsum}

\newcommand{\tabularxmulticolumncentered}[3] 
    {\multicolumn{#1}
                 {>{\hsize=\dimexpr#1\hsize+#1\tabcolsep+\arrayrulewidth\relax}#2}
                 {#3}}

\algnewcommand\algorithmicparams{\textbf{Parameters:}}
\algnewcommand\Params{\item[\algorithmicparams]}

\usepackage{tikz}
\usepackage{pgfplots}
\pgfplotsset{compat=1.16}
\usetikzlibrary{backgrounds,arrows,automata,shapes,positioning,calc,through}
\pgfdeclarelayer{background}
\pgfdeclarelayer{foreground}
\pgfsetlayers{background,main,foreground}

\usetikzlibrary{fit}

\tikzset{
  highlight/.style={
    circle,
    draw=blue, very thick,
    minimum height=2.5em,
    text centered,
    inner sep = 0pt,
  },
  highlight2/.style={
    circle,
    draw=red, very thick,
    minimum height=5.8em,
    text centered,
    inner sep = 0pt,
  },
  state/.style={
    rectangle,
    draw=black, very thick,
    minimum height=1.0em,
    text centered,
  },
  final_state/.style={
    rectangle,
    rounded corners,
    draw=black, very thick,
    minimum height=2em,
    text centered,
    dashed
  },
  initial_state/.style={
    rectangle,
    double=white,
    double distance=1pt,
    inner sep=2pt,
    draw=black, very thick,
    minimum height=2em,
    text centered,
  },
  point/.style={
    circle,
    inner sep=0pt,
    minimum size=3pt,
    fill=red
  }
}

\usepackage{pstricks}
\usepackage{tikz-3dplot}

\makeatletter
\let\NAT@parse\undefined
\makeatother
\usepackage{hyperref}

\newcommand{\bl}[1]{{{#1}}}

\renewcommand\vec{\bm}
\newcommand\mat{\mathbf}

\newcommand{\frames}[3]{{\prescript{{#1}}{{#2}}{#3}}}
\usepackage{cancel}

\title{\LARGE \bf Drones Guiding Drones: Cooperative Navigation of a Less-Equipped Micro Aerial Vehicle in Cluttered Environments}

\author{Václav Pritzl$^{*}$, Matouš Vrba$^{*}$, Yurii Stasinchuk$^{*}$, Vít Krátký$^{*}$, Jiří Horyna$^{*}$, Petr Štěpán$^{*}$, and Martin Saska$^{*}$
\thanks{
This work was supported by CTU grant no SGS23/177/OHK3/3T/13, by the Czech Science Foundation (GAČR) under research project No. 23-07517S, and by the European Union under the project Robotics and advanced industrial production (reg. no. CZ.02.01.01/00/22\_008/0004590).}
\thanks{$^{*}$The authors are with the Multi-robot Systems Group, Department of Cybernetics, Faculty of Electrical Engineering,
        Czech Technical University in Prague, Czech Republic
        {\tt\small \{vaclav.pritzl, matous.vrba, yurii.stasinchuk, vit.kratky, jiri.horyna, petr.stepan, martin.saska\}@fel.cvut.cz}}%
}

\newcommand{\PREPRINTYEAR}{2024}
\newcommand{\PUBLISHEDIN}{IEEE}
\newcommand{\PUBLISHEDINSHORT}{IROS 2024}
\newcommand{\DOI}{10.1109/IROS58592.2024.10802770} 

\usepackage[placement=top,vshift=-7]{background}
\SetBgScale{1.0}
\SetBgContents{\PUBLISHEDINSHORT. ACCEPTED VERSION - DO NOT DISTRIBUTE. \href{https://doi.org/\DOI}{DOI \DOI}}
\SetBgColor{black}
\SetBgAngle{0}
\SetBgOpacity{1.0}

\begin{document}

\thispagestyle{empty}
\onecolumn
{
  \topskip0pt
  \vspace*{\fill}
  \centering
  \LARGE{%
    \copyright{} \PREPRINTYEAR~\PUBLISHEDIN\\\vspace{1cm}
    Personal use of this material is permitted.
    Permission from IEEE~must be obtained for all other uses, in any current or future media, including reprinting or republishing this material for advertising or promotional purposes, creating new collective works, for resale or redistribution to servers or lists, or reuse of any copyrighted component of this work in other works.}
    \vspace*{\fill}
}
\NoBgThispage
\twocolumn          	
\BgThispage

\bstctlcite{IEEEexample:BSTcontrol}


\maketitle
\thispagestyle{empty}
\pagestyle{empty}

\begin{abstract}
  Reliable deployment of \acp{UAV} in cluttered unknown environments requires accurate sensors for \ac{GNSS}-denied localization and obstacle avoidance.
  Such a requirement limits the usage of cheap and micro-scale vehicles with constrained payload capacity if industrial-grade reliability and precision are required.
  This paper investigates the possibility of offloading the necessity to carry heavy sensors to another member of the \ac{UAV} team while preserving the desired capability of the smaller robot intended for exploring narrow passages.
  A novel cooperative guidance framework offloading the sensing requirements from a minimalistic secondary \ac{UAV} to a superior primary \ac{UAV} is proposed.
  The primary \ac{UAV} constructs a dense occupancy map of the environment and plans collision-free paths for both \acp{UAV} to ensure reaching the desired secondary \ac{UAV}'s goals even in areas not accessible by the bigger robot.
  The primary \ac{UAV} guides the secondary \ac{UAV} to follow the planned path while tracking the \ac{UAV} using \ac{lidar}-based relative localization.
  The proposed approach was verified in real-world experiments with a heterogeneous team of a 3D \ac{lidar}-equipped primary \ac{UAV} and a micro-scale camera-equipped secondary \ac{UAV} moving autonomously through unknown cluttered \ac{GNSS}-denied environments with the proposed framework running fully on board the \acp{UAV}.

\end{abstract}


\section*{Multimedia Attachment}
\noindent\url{https://mrs.felk.cvut.cz/drones-guiding}


\section{INTRODUCTION}
\label{sec:introduction}
\vspace{-0.1em}

\acresetall

The ability to accurately perceive the surrounding obstacles and localize in an onboard-built map is crucial for aerial robots operating in unknown cluttered environments.
However, such capabilities may require the presence of heavy sensors on board the \acp{UAV}, significantly increasing their dimensions.
Minimizing such hardware requirements is critical when the \acp{UAV} need to be as small as possible to be able to operate in narrow passages.

In many applications, having some micro \acp{UAV} with only application-specific payload is desirable.
In cooperative sensing tasks, the aim is to distribute \acp{UAV} with specific sensors to target positions to detect the sought phenomenon, such as a gas source~\cite{duisterhofSniffyBugFully2021} or radiation~\cite{stibinger2020ral}.
In a confined indoor environment, small \acp{UAV} with communication modules may be used to build a mesh communication network, as in the breadcrumb-deploying communication solutions of the DARPA SubT challenge~\cite{petrlikUAVsSurfaceCooperative2022}.
In \ac{GNSS}-denied conditions, \acp{UAV} carrying localization anchors can be distributed to specific positions in the environment to provide an external localization system for other robots~\cite{natterIncrementallyDeployedSwarm2022}.
In sensory degraded environments, the minimalistic \acp{UAV} can serve as landmarks for improving the localization performance of the rest of the robot team~\cite{spasojevicActiveCollaborativeLocalization2023a}.
All these tasks benefit from the ability to guide cheap miniature \acp{UAV} with only application-specific hardware to target positions or to places untraversable by the larger \acp{UAV}, either guiding them all the way or through areas too challenging for their own onboard sensors.

\begin{figure}[t]
  \centering
  \begin{tikzpicture}
    \node[anchor=north west,inner sep=0,draw=black] (a) at (0, 0)
    {
      \includegraphics[width=0.495\linewidth, trim=1cm 0.5cm 1.86cm 0.5cm, clip=true]{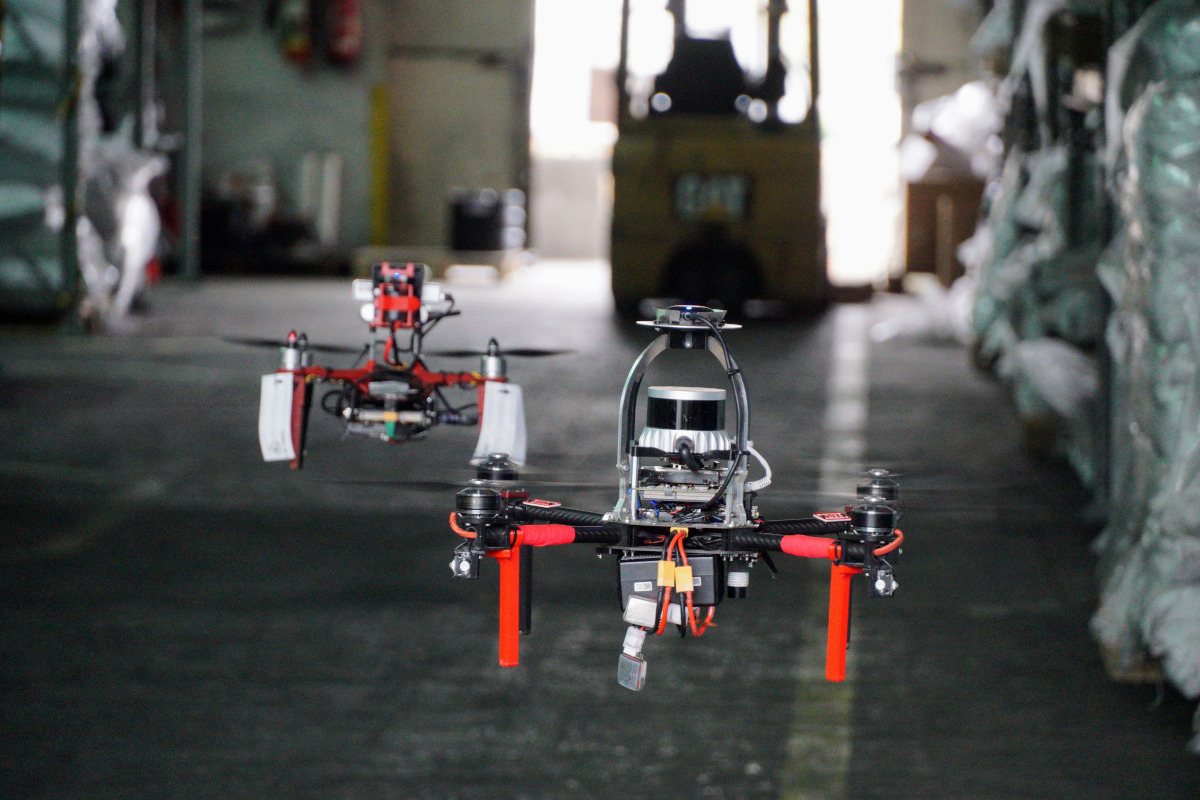}
    };
    \node[fill=white,draw=black,text=black, anchor=south west] at (a.south west) {\footnotesize (a)};

    \node[anchor=north west,inner sep=0,draw=black] (b) at (4.35cm, 0cm)
    {
      \includegraphics[width=0.495\linewidth, trim=0.7cm 1.3cm 1.145cm 0.28cm, clip=true]{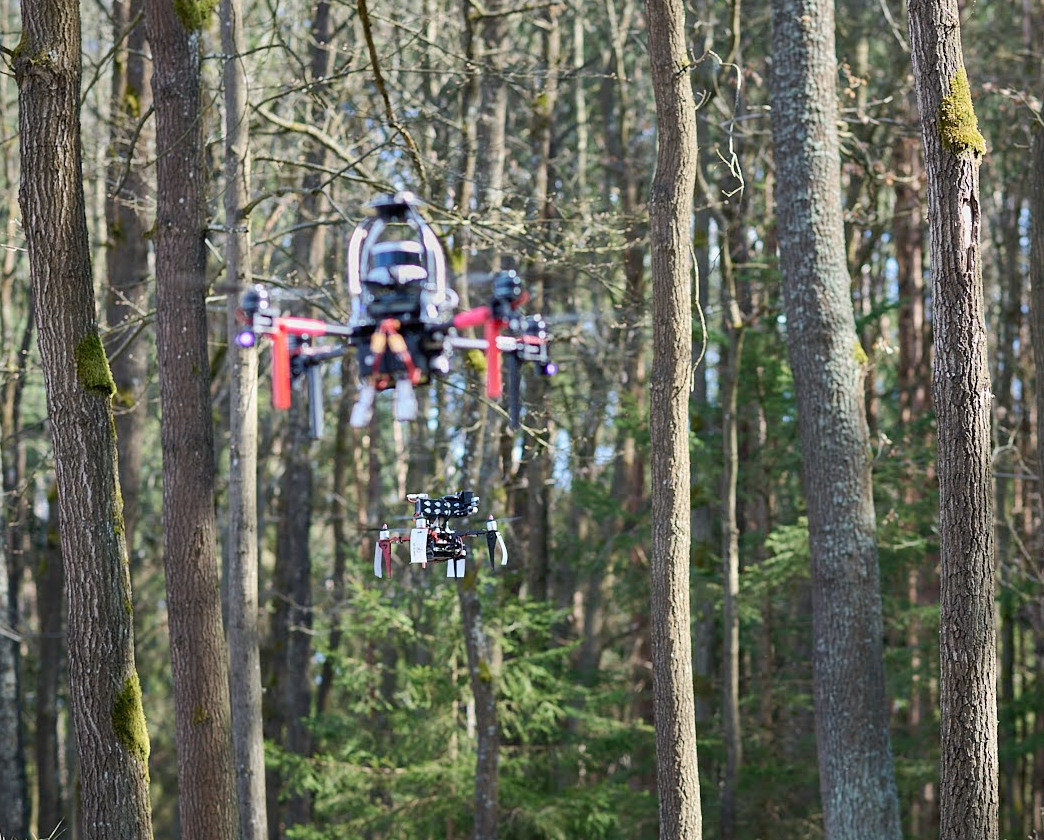}
    };
    \node[fill=white,draw=black,text=black, anchor=south west] at (b.south west) {\footnotesize (b)};

    \node[fill=white, rounded corners,opacity=0.9] (label1) at (3.2, -0.9) {
      \scriptsize
      \color{black}{Primary UAV}
    };
    \node[fill=white, rounded corners,opacity=0.9] (label2) at (1.2, -0.7) {
      \scriptsize
      \color{black}{Secondary UAV}
    };

  \end{tikzpicture}
  \caption{The \ac{lidar}-equipped primary \ac{UAV} guides the secondary camera-equipped \ac{UAV} through (a) narrow passages in an industrial complex, (b) a cluttered forest environment.%
  }
  \label{fig:motivation}
  \vspace{-1.3em}
\end{figure}
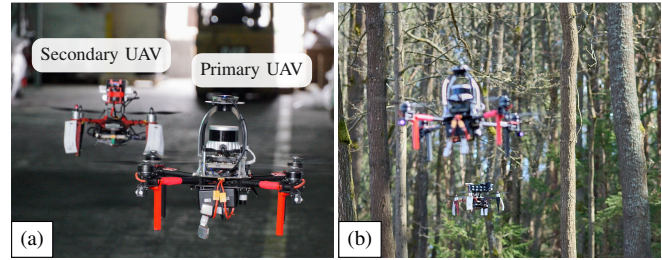

In this paper, we focus on offloading the sensing capability from the less-capable \acp{UAV} to a more-capable guiding \ac{UAV} with superior sensors and processing power.
Specifically, we deal with the case of a 3D \ac{lidar}-equipped primary \ac{UAV} guiding a smaller camera-equipped secondary \ac{UAV} (see Fig.~\ref{fig:motivation}).
3D \acp{lidar} are precise and robust to various environmental conditions but are generally heavy.
Combining \ac{lidar}-based and visual-based navigation reaps the benefits of both approaches while mitigating their drawbacks.

\subsection{Problem statement}
\label{sec:problem_statement}
\vspace{-0.1em}

We tackle the problem of cooperative flight of two diverse \acp{UAV} in an unknown \ac{GNSS}-denied cluttered environment.
The \ac{PUAV} carries a 3D \ac{lidar} sensor.
The \ac{SUAV} carries a visual camera for local self-localization only and has insufficient sensing capabilities for precise global localization and obstacle detection.
Both \acp{UAV} are equipped with an onboard computer, an \ac{IMU}, an embedded attitude controller, and a wireless communication module.
We assume that the 3D \ac{lidar} provides sufficient data for precise localization with low long-term drift w.r.t. the world reference frame, while the visual camera data is sufficient for short-term localization only, exhibiting larger drift than the \ac{lidar}-based localization.
Relying on the \ac{lidar} sensor for self-localization of the \ac{PUAV} and relative localization between the \acp{UAV} enables precise localization for the entire team.
The \acp{UAV} are capable of mutual communication with the \ac{PUAV} sending planned paths to the \ac{SUAV} and the \ac{SUAV} transmitting its local odometry data to the \ac{PUAV}.
All software runs on board the \acp{UAV} with no external computational resources.
Both \acp{UAV} use only their onboard sensors for localization, and no external localization system is available.

We denote vectors as bold lowercase letters, matrices as bold upright uppercase letters, and frames of reference as uppercase italics letters.
Let $\frames{B}{A}{\mat{T}} \in SE(3)$ be the transformation matrix describing the transformation from frame ${A}$ to frame ${B}$.
Let $\frames{A}{}{\vec{x}}$ be a 3D position vector in frame ${A}$.
We denote sets and sequences by uppercase calligraphic letters.
Let $\frames{A}{}{\mathcal{P}_\mathrm{U}}$ be a sequence of \ac{UAV} reference poses  $(\frames{A}{}{\vec{x}}_i, \frames{A}{}{\theta}_i)$, consisting of position $\frames{A}{}{\vec{x}}_i \in \mathbb{R}^3$ and heading $\frames{A}{}{\theta}_i \in [-\pi,\pi]$, for \ac{UAV} $\mathrm{U}$ in reference frame $A$, thus forming a path with a specific heading assigned to each position.

\begin{figure}
\centering
  \includegraphics[width=0.9\linewidth, trim=0.0cm 0.0cm 0.0cm 0.0cm, clip=true]{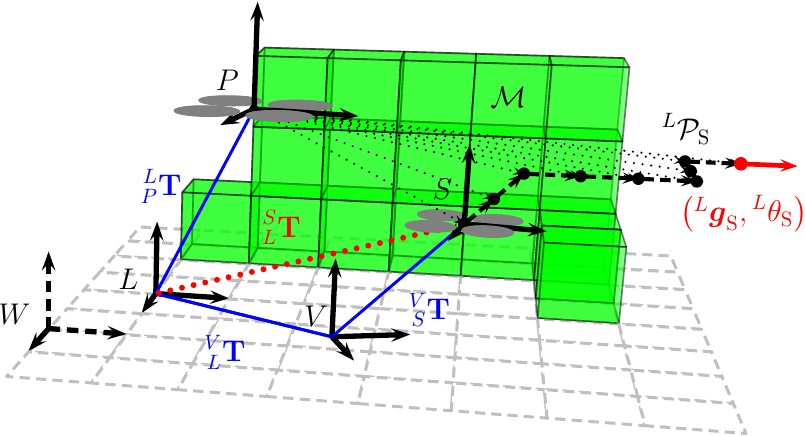}
  \caption{The \ac{PUAV} with body frame ${P}$ is localized in its local frame ${L}$, builds dense occupancy map $\mathcal{M}$, and plans collision-free paths for both \acp{UAV}.
  The \ac{SUAV} with body frame ${S}$ is localized in its local frame ${V}$. ${W}$ denotes the fixed world frame.
  All the reference frames are gravity-aligned.
  The \ac{PUAV} periodically guides the \ac{SUAV} to follow the planned path $\frames{L}{}{\mathcal{P}_\mathrm{S}}$.
  Black dotted lines mark the line of sight between the \ac{PUAV} position and the \ac{SUAV} waypoints.}
\label{fig:drones_tikz}
  \vspace{-1.3em}
\end{figure}

The addressed problem, depicted in Fig.~\ref{fig:drones_tikz}, is to guide the \ac{SUAV} to a goal pose $\left(\frames{L}{}{\vec{g}}_\mathrm{S}, \frames{L}{}{\theta_\mathrm{S}}\right)$, consisting of position $\frames{L}{}{\vec{g}}_\mathrm{S} \in \mathbb{R}^3$ and heading $\frames{L}{}{\theta_\mathrm{S}} \in [-\pi,\pi]$.
Specifically, the task is to plan collision-free paths $\frames{L}{}{\mathcal{P}_\mathrm{S}}$ and $\frames{L}{}{\mathcal{P}_\mathrm{P}}$ based on the 3D occupancy map $\mathcal{M}$ built by the \ac{PUAV} and guide the \ac{SUAV} along the planned path $\frames{L}{}{\mathcal{P}_\mathrm{S}}$, while $\frames{L}{}{\mathcal{P}_\mathrm{P}}$ is planned such that the \ac{PUAV} tracks the \ac{SUAV} throughout the environment.

\subsection{Related work}\label{sec:SoA}
\vspace{-0.1em}


Utilizing heterogeneous multi-robot teams consisting of more-capable robots guiding less-capable robots has been proposed for use with \acp{UGV} in the past~\cite{parkerTightlycoupledNavigationAssistance2004, huangLocalizationFollowtheleaderControl2006, howardExperimentsLargeHeterogeneous2006, hofmeisterCooperativeVisualMapping2011}.
Works in the area of \ac{UGV}-\ac{UAV} cooperation have focused on the cooperation of a \ac{UGV} with a \ac{UAV} acting as an ``eye in the sky"~\cite{hsiehAdaptiveTeamsAutonomous2007}.
The \ac{UAV} can map the surroundings of the \ac{UGV} and plan a collision-free path for the \ac{UGV} to follow~\cite{zhangFastActiveAerial2022, muegglerAerialguidedNavigationGround2014a, delmericoActiveAutonomousAerial2017, fankhauserCollaborativeNavigationFlying2016}.
The path planning can be performed directly on the image data~\cite{lakasFrameworkCooperativeUAVUGV2018, changDroneAidedPathPlanning2022} or as a \ac{POMDP}~\cite{chenMotionPlanningHeterogeneous2020} without global map construction.
In~\cite{shenCollaborativeAirgroundTarget2017}, the \ac{UAV} sent only target positions to the \ac{UGV}, which performed motion planning on its own sensory data.
In~\cite{chenGANbasedActiveTerrain2019}, a terrain map was constructed on a ground station using a learning-based approach and sent to the \ac{UGV}, which used it for path planning.
In~\cite{mikiUAVUGVAutonomous2019}, the \ac{UAV} shared its voxel map of the environment with a \ac{UGV}, which converted the voxel map to an elevation map and utilized it for path planning.
In~\cite{leeAerialOnlineMapping2023}, a \ac{UAV} detected a \ac{UGV} along with obstacles on the ground and shared the data with the \ac{UGV} to aid the \ac{UGV}'s path planning.
In~\cite{liColAGCollaborativeAirGround2023}, a framework for a single \ac{UAV} guiding a group of blind \acp{UGV} among obstacles was proposed.
In~\cite{hsiehAdaptiveTeamsAutonomous2007, zhangFastActiveAerial2022, muegglerAerialguidedNavigationGround2014a, delmericoActiveAutonomousAerial2017, changDroneAidedPathPlanning2022, chenMotionPlanningHeterogeneous2020, shenCollaborativeAirgroundTarget2017, chenGANbasedActiveTerrain2019, mikiUAVUGVAutonomous2019, leeAerialOnlineMapping2023}, the authors either assume apriori alignment of the \ac{UAV}-\ac{UGV} reference frames or utilize camera-based relative localization.
However, camera-based localization may suffer from inaccurate distance estimation of the detected teammate or obstacles, as noted in~\cite{leeAerialOnlineMapping2023}.
While its accuracy is sufficient for guiding a \ac{UGV} among obstacles on the ground, guiding a \ac{UAV} through 3D space requires more accurate localization.
In~\cite{lakasFrameworkCooperativeUAVUGV2018}, an \ac{UWB}-based localization system was used, but such a system requires apriori-placed anchors in the environment.
In~\cite{fankhauserCollaborativeNavigationFlying2016}, matching visual landmarks in shared maps was employed, but sharing map data requires significant communication bandwidth.
The authors of~\cite{liColAGCollaborativeAirGround2023} utilized relative pose estimation from detections of infrared LED markers, \ac{UWB}, and \ac{IMU} data, but such an approach requires a specialized localization module on each robot.
All of the aforementioned \ac{UAV}-\ac{UGV} cooperation approaches assume that the \ac{UAV} is moving above the obstacles and does not need to solve obstacle avoidance nor maintaining \ac{LOS} of the teammates w.r.t. the obstacles.

In~\cite{masnaviVACNAVisibilityAwareCooperative2023a}, a trajectory optimization method for a 3D \ac{lidar}-carrying \ac{UGV} guiding a camera-equipped \ac{UAV} while keeping visibility between obstacles was proposed.
The approach was evaluated mainly in simulations and localization details were not provided.
The obstacle avoidance and occlusion mitigation were treated as 2D problems, while we solve the obstacle avoidance in 3D and occlusion mitigation in 2D.
Finally, in a \ac{UAV}-{UGV} system, the \ac{UAV} can safely fly over a \ac{UGV}.
In a multi-\ac{UAV} system, the \acp{UAV} need to maintain sufficient altitude separation to avoid propeller downwash.

Regarding multi-\ac{UAV} cooperation, the closest methods~\cite{jangNavigationAssistantPathPlanning2020a, wangVisibilityawareTrajectoryOptimization2021, leeTargetvisiblePolynomialTrajectory2021, yinDecentralizedSwarmTrajectory2023} to our approach used a leader-follower scheme with one \ac{UAV} moving through the environment and the other \acp{UAV} tracking the leader.
In~\cite{jangNavigationAssistantPathPlanning2020a}, the navigation of a main-agent was aided by a monocular camera-equipped sub-agent.
The approach did not consider obstacles and was evaluated in simulations only.
A leader-follower scheme was applied in trajectory generation methods for tracking a UAV among obstacles using a camera~\cite{wangVisibilityawareTrajectoryOptimization2021, leeTargetvisiblePolynomialTrajectory2021}, and in a decentralized system for tracking a target UAV by a UAV swarm among obstacles, using LiDAR-based relative localization~\cite{yinDecentralizedSwarmTrajectory2023}.

The aforementioned leader-follower-based studies~\cite{jangNavigationAssistantPathPlanning2020a, wangVisibilityawareTrajectoryOptimization2021, leeTargetvisiblePolynomialTrajectory2021, yinDecentralizedSwarmTrajectory2023} dealt with \acp{UAV} that are equal in size.
They would fail in mixed-size \ac{UAV} team scenarios requiring a miniature \ac{UAV} with limited or absent obstacle-sensing capability to pass through a narrow gap untraversable for the leader.
In such scenarios, which match inspection and data-gathering tasks in hard-to-reach areas, the obstacle avoidance and visibility-maintaining constraints conflict as the leading \ac{UAV} cannot directly lead a secondary \ac{UAV} through narrow openings.
Thus, the leader must guide the \ac{UAV} from a distance using communication and precise relative localization.

To obtain high-accuracy relative localization, necessary for our guiding approach, we utilize \ac{lidar}-based relative localization, similarly as in~\cite{yinDecentralizedSwarmTrajectory2023}, but focus on a heterogeneous \ac{UAV} team where only the \ac{PUAV} is carrying a 3D \ac{lidar} while the minimalistic \ac{SUAV} carries a camera providing data for \ac{VIO}.
We build upon our previous works on the use of \ac{lidar}-based \ac{UAV} detections~\cite{pritzl2022icuas} and fusion of these detections with \ac{VIO}~\cite{pritzlFusionVisualInertialOdometry2023} for guiding a cooperating \ac{UAV}.
To the best of our knowledge, our approach marks the first deployment of a heterogeneous \ac{UAV} team offloading obstacle detection and motion planning from one \ac{UAV} to another.
The contributions of this paper are summarized as:
\begin{itemize}
  \item A novel approach for flight of a heterogeneous \ac{UAV} team in cluttered environments, offloading occupancy mapping and path planning requirements to a more-capable \ac{PUAV} while preserving the obstacle avoidance capabilities and precise localization of the less-capable \ac{SUAV}.
    The framework ensures accurate guiding of the \ac{SUAV} with minimal communication requirements.
  \item A unique cooperative relative-localization-aware path planning method for a mixed-size \ac{UAV} team.
    Collision-free paths are planned for the \ac{SUAV}, while the \ac{PUAV} motion is planned to maximize \ac{LOS} visibility between the \acp{UAV} before a \ac{LOS} break, which may be unavoidable due to their different sizes.
\end{itemize}
The proposed methodology was evaluated in realistic simulations as well as in demanding real-world deployments while running on board the \acp{UAV} with no external computational resources and no external localization system.


\section{MULTI-UAV PLANNING AND GUIDANCE}
\label{sec:system_model}
\vspace{-0.1em}

\begin{figure*}
  \centering
  \includegraphics[width=0.95\linewidth, trim=0.0cm 0.0cm 0.0cm 0.0cm, clip=true]{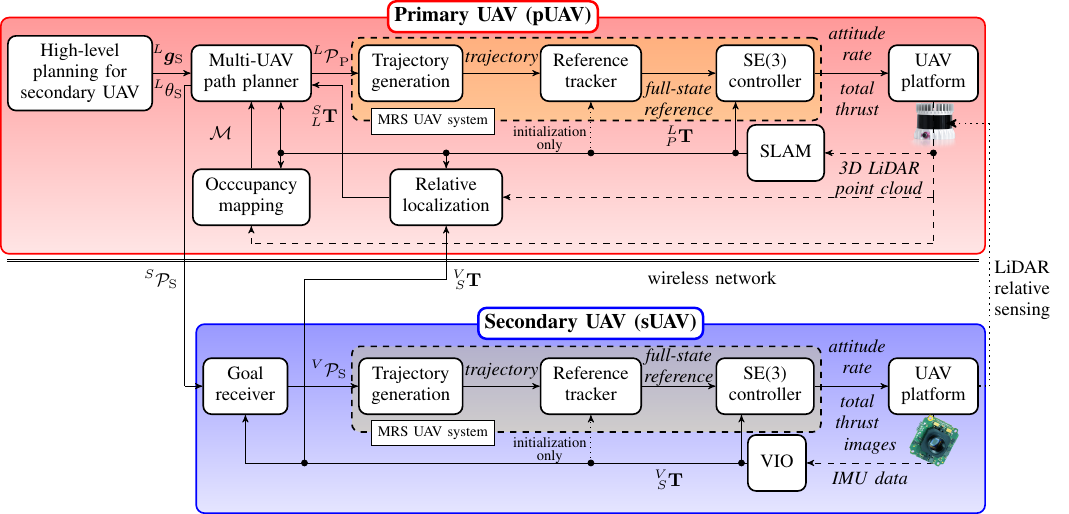}
  \caption{The \ac{PUAV} maps the surrounding environment using 3D \ac{lidar} data, performs estimation of relative pose of the \ac{SUAV} from \ac{lidar} detections and received \ac{VIO} data, and guides the \ac{SUAV} to desired poses. The position control pipeline in the frame of each respective self-localization method is provided by the MRS UAV System~\cite{bacaMRSUAVSystem2021}.}
  \label{fig:diagram}
  \vspace{-1.3em}
\end{figure*}

A high-level diagram of the proposed multi-\ac{UAV} framework is shown in Fig.~\ref{fig:diagram}.
The \textit{Multi-UAV path planner} module receives high-level \ac{SUAV} goals $\left(\frames{L}{}{\vec{g}}_\mathrm{S}, \frames{L}{}{\theta_\mathrm{S}}\right)$ and constructs collision-free paths for both the \ac{SUAV} and the \ac{PUAV} itself.
The \ac{SUAV} path is periodically transformed to the \ac{SUAV} body frame ${S}$ using the relative localization data and transmitted to the \ac{SUAV} over the wireless network.
During the guiding process, the \ac{PUAV}-\ac{SUAV} pair essentially forms a closed loop controlling the \ac{SUAV} pose in the local reference frame ${L}$.
The \ac{PUAV} is localized using a 3D \ac{lidar}-based \ac{SLAM} and performs dense occupancy mapping using the Octomap~\cite{hornungOctoMapEfficientProbabilistic2013} approach, producing an occupancy map $\mathcal{M}$ of the environment.
To decrease the processing requirements, the map $\mathcal{M}$ represents a limited region of the environment, centered at the current \ac{PUAV} position.
The \ac{SUAV} uses \ac{VIO} for self-localization.
Generating time-parametrized trajectories from planned paths and position control in the frames of the respective self-localization methods is provided by the MRS UAV System\footnote{\url{https://github.com/ctu-mrs/mrs_uav_system}}~\cite{bacaMRSUAVSystem2021} running on each \ac{UAV}.
The system times of the \acp{UAV} are synchronized over the wireless network using the \ac{NTP}.

\subsection{Proposed multi-UAV path planner}
\vspace{-0.1em}

\begin{algorithm}[!ht]
  \small
  \caption{Multi-UAV path planning (state = \textit{PLANNING})}
    \label{alg:planning}
    \begin{algorithmic}[1] 
      \Require Map $\mathcal{M}$, \ac{PUAV} pose $\left(\frames{L}{}{\vec{x}_\mathrm{P}}, \frames{L}{}{\phi_\mathrm{P}}\right)$, \ac{SUAV} pose $\left(\frames{L}{}{\vec{x}_\mathrm{S}}, \frames{L}{}{\phi_\mathrm{S}}\right)$, \ac{SUAV} goal pose $\left(\frames{L}{}{\vec{g}_\mathrm{S}}, \frames{L}{}{\theta_\mathrm{S}}\right)$
      \Ensure Paths $\frames{L}{}{\mathcal{P}_\mathrm{P}}, \frames{L}{}{\mathcal{P}_\mathrm{S}}$ sent to the \acp{UAV}' control pipelines
      \Params \ac{PUAV} obstacle width $w_\mathrm{P}$ and height $h_\mathrm{P}$, \ac{SUAV} obstacle width $w_\mathrm{S}$ and height $h_\mathrm{S}$, \ac{PUAV} minimum obstacle distance $d_\mathrm{P}$, \ac{SUAV} minimum obstacle distance $d_\mathrm{S}$

      \If{$\left|\left|\frames{L}{}{\vec{x}}_\mathrm{S} - \frames{L}{}{\vec{g}}_\mathrm{S}\right|\right|_2 < \mathcal{M}.resolution$}
        \State \Call{change\_state}{\textit{GOAL REACHED}}
        \State \textbf{return}
      \EndIf
      \State $\mathcal{M}_\mathrm{S} \gets \Call{copy\_map\_and\_add\_obstacle}{\mathcal{M}$, $\frames{L}{}{\vec{x}}_\mathrm{P}$, $w_\mathrm{P}$, $h_\mathrm{P}}$
      \State $\frames{L}{}{\mathcal{P}_\mathrm{S}} \gets \Call{find\_path}{\mathcal{M}_\mathrm{S}, \frames{L}{}{\vec{x}_\mathrm{S}}, \frames{L}{}{\phi_\mathrm{S}}, \frames{L}{}{\vec{g}_\mathrm{S}}, \frames{L}{}{\theta_\mathrm{S}},d_\mathrm{S}}$ \Comment{Using A* planner}
      \If{$\frames{L}{}{\mathcal{P}_\mathrm{S}} = \varnothing$}
        \State \Call{change\_state}{\textit{FAILURE}}
        \State \textbf{return}
      \EndIf
      \State $\frames{L}{}{\vec{g}_\mathrm{P}} \gets \Call{find\_guiding\_viewpoint}{\mathcal{M}, \frames{L}{}{\mathcal{P}_\mathrm{S}}, \frames{L}{}{\vec{x}_\mathrm{P}}}$ \Comment{Alg.~2}
      \If{$\frames{L}{}{\vec{g}_\mathrm{P}} = \varnothing$}
        \State \Call{change\_state}{\textit{FAILURE}}
        \State \textbf{return}
      \EndIf
      \If{$\left|\left|\frames{L}{}{\vec{x}}_\mathrm{P} - \frames{L}{}{\vec{g}}_\mathrm{P}\right|\right|_2 < \mathcal{M}.resolution$}
        \State \Call{change\_state}{\textit{SECONDARY MOVING}, $\frames{L}{}{\mathcal{P}_\mathrm{S}}$}
        \State \textbf{return}
      \EndIf
      \State $\mathcal{M}_\mathrm{P} \gets \Call{copy\_map\_and\_add\_obstacle}{\mathcal{M}$, $\frames{L}{}{\vec{x}}_\mathrm{S}$, $w_\mathrm{S}$, $h_\mathrm{S}}$
      \State $\frames{L}{}{\mathcal{P}_\mathrm{P}} \gets \Call{find\_path}{\mathcal{M}_\mathrm{P}, \frames{L}{}{\vec{x}_\mathrm{P}}, \frames{L}{}{\phi_\mathrm{P}}, \frames{L}{}{\vec{g}_\mathrm{P}}, \frames{L}{}{\theta_\mathrm{P}},d_\mathrm{P}}$ \Comment{Using A* planner}
      \If{$\frames{L}{}{\mathcal{P}_\mathrm{P}} = \varnothing$}
        \State \Call{change\_state}{\textit{FAILURE}}
        \State \textbf{return}
      \EndIf
      \State \Call{change\_state}{\textit{PRIMARY MOVING}, $\frames{L}{}{\mathcal{P}_\mathrm{P}}$}
        \State \textbf{return}

    \end{algorithmic}
\end{algorithm}

\begin{algorithm}[!ht]
  \small
  \caption{Finding a guiding viewpoint}
    \label{alg:vis_planning}
    \begin{algorithmic}[1] 
      \Require Map $\mathcal{M}$, \ac{PUAV} position $\frames{L}{}{\vec{x}}_\mathrm{P}$, \ac{SUAV} path $\frames{L}{}{\mathcal{P}_\mathrm{S}}$
      \Ensure Goal $\frames{L}{}{\vec{g}}_\mathrm{P}$ for the \ac{PUAV}
      \Params raycasting sample count $n_\mathrm{samples}$, raycasting max. ray length $d_\mathrm{ray}$, min. safe distance $d_\mathrm{buffer}$ from the \ac{SUAV} path, \ac{PUAV} min. obstacle distance $d_\mathrm{P}$

      \LineComment{Construct a safety buffer around the \ac{UAV} path}
      \State $\mathcal{B} \gets \Call{buffer}{\frames{L}{}{\mathcal{P}_\mathrm{S}},d_\mathrm{buffer}}$ \Comment{Single non-convex polygon}
      \LineComment{Get a visibility region for each point in the path}
      \State $\mathcal{V_\mathrm{all}} \gets \varnothing$ \Comment{Set of multi-polygons}
      \For{$(\vec{p}_i,\theta_i) \in \frames{L}{}{\mathcal{P}_\mathrm{S}}$}
        \State $\mathcal{V}_i \gets \varnothing$ \Comment{Single non-convex polygon}

        \For{$\alpha \in \left( k \frac{2\pi}{n_\mathrm{samples}} \right)_{k=0}^{n_\mathrm{samples}}$}

          \State $\vec{r}_\mathrm{end} \gets \vec{p}_i + [d_\mathrm{ray}\cos\alpha,d_\mathrm{ray}\sin\alpha,0]^\mathrm{T}$
          \LineComment{get the first intersection with obstacle or $\vec{r}_\mathrm{end}$}
          \State $\vec{r}_\mathrm{int} \gets \Call{ray\_obstacle\_intersect}{\mathcal{M}, \vec{p}_i,\vec{r}_\mathrm{end}}$
          \State $\mathcal{V}_i.\mathrm{add}(\vec{r}_\mathrm{int})$

        \EndFor
      \State $\mathcal{V}_\mathrm{all}.\mathrm{add}(\mathcal{V}_i \setminus \mathcal{B})$

      \EndFor

      \LineComment{Generate safe space region}
      \State $\mathcal{S} \gets \varnothing$ \Comment{Multi-polygon (collection of polygons)}
      \State $ r \gets \mathcal{M}.resolution$
      \For{$v \in \mathcal{M}[v_z = \frames{L}{}{x_{\mathrm{P}z}}]$} \Comment{$v$ is a single voxel}
          \If{$\Call{obs\_dist}{v} > d_\mathrm{P}~\algand~v\in\bigcup\mathcal{V}_\mathrm{all}$}

            \State $\mathcal{S} \gets \mathcal{S} \cup \Call{polygon}{v_x \pm \frac{r}{2},v_y\pm \frac{r}{2}}$ \Comment{Do union with safe space region, adjacent polygons are joined}

          \EndIf
        \EndFor
      \State $\mathcal{S}_\mathrm{closest} \gets \arg\min_{\mathcal{S}_i\in\mathcal{S}} \Call{distance}{\frames{L}{}{\vec{x}_\mathrm{P}},\mathcal{S}_i}$
      
      \LineComment{Find intersection of safe region and the visibility regions}
      \State $\mathcal{I}_\mathrm{all} \gets \mathcal{S}_\mathrm{closest}$
      \State $i_\mathrm{found} \gets \mathbf{false}$
      \For{$\mathcal{V}_i \in \mathcal{V}_\mathrm{all}$} \Comment{Ordered same as $\frames{L}{}{\mathcal{P}}_\mathrm{S}$, start from $\mathcal{V}_0$}
          \State $\mathcal{I}_\mathrm{new} \gets \mathcal{I}_\mathrm{all} \cap \mathcal{V}_i$ \Comment{Intersection is a multi-polygon}
          
        \If{$\mathcal{I}_\mathrm{new} = \varnothing$} \Comment{No more intersections found}
          \State \textbf{break}
        \Else
          \State $\mathcal{I}_\mathrm{all} \gets \mathcal{I}_\mathrm{new}$
          \State $i_\mathrm{found} \gets \mathbf{true}$
        \EndIf
      \EndFor

      \If{$i_\mathrm{found}$}
        \LineComment{Select goal as the closest pole of inaccessibility}
        \State \textbf{return} $\textup{POI}\left(\arg\min_{\mathcal{I}_i \in \mathcal{I}_\mathrm{all}} \left|\left| \Call{POI}{\mathcal{I}_i} - \frames{L}{}{\vec{x}}_\mathrm{P}\right|\right|_2\right)$
        \Else
          \State \textbf{return} $\varnothing$
      \EndIf

    \end{algorithmic}
\end{algorithm}

The \textit{Multi-UAV path planner} module consists of a \ac{FSM} transferring between the \textit{IDLE}, \textit{PLANNING}, \textit{PRIMARY MOVING}, \textit{SECONDARY MOVING}, \textit{GOAL REACHED}, and \textit{FAILURE} states.
When the module receives a new goal $\left(\frames{L}{}{\vec{g}}_\mathrm{S}, \frames{L}{}{\theta_\mathrm{S}}\right)$ for the \ac{SUAV}, the \ac{FSM} transfers to the \textit{PLANNING} state and constructs collision-free paths $\frames{L}{}{\mathcal{P}_\mathrm{P}},\frames{L}{}{\mathcal{P}_\mathrm{S}}$ for both \acp{UAV}.
In the \textit{PRIMARY MOVING} state, the \ac{PUAV} moves along its path to a viewpoint $\frames{L}{}{\vec{g}_\mathrm{P}}$, where it can observe and guide the \ac{SUAV} along $\frames{L}{}{\mathcal{P}_\mathrm{S}}$.
After reaching the viewpoint, the \ac{FSM} transfers to the \textit{SECONDARY MOVING} state.
In the \textit{SECONDARY MOVING} state, the \ac{PUAV} guides the \ac{SUAV} to follow the planned collision-free path $\frames{L}{}{\mathcal{P}_\mathrm{S}}$.
When the goal is reached, the \ac{FSM} transfers to the \textit{GOAL REACHED} state.

The path planning process proceeds according to Alg.~\ref{alg:planning}.
The planner takes the current occupancy map $\mathcal{M}$ and inserts an occupied region at the current position $\frames{L}{}{\vec{x}}_\mathrm{P}$ of the \ac{PUAV} with safety margin above and below to avoid the downwash effect.
A collision-free path $\frames{L}{}{\mathcal{P}_\mathrm{S}}$ for the \ac{SUAV} is planned using an A$^*$ planner.
We have utilized the A$^*$ planning algorithm with iterative path post-processing that we designed for the DARPA SubT challenge, described in~\cite{kratky2021exploration, petrlikUAVsSurfaceCooperative2022}.
Based on the resulting path, a goal viewpoint $\frames{L}{}{\vec{g}_\mathrm{P}}$ for the \ac{PUAV} is selected to maximize the uninterrupted sequence of waypoints, beginning at $\frames{L}{}{\vec{x}}_\mathrm{S}$, visible by the \ac{PUAV} (see sec.~\ref{sec:viewpoint} for details).
Occupied space at the position $\frames{L}{}{\vec{x}}_\mathrm{S}$ of the \ac{SUAV} is inserted to the map and the A$^*$ planner constructs a collision-free path $\frames{L}{}{\mathcal{P}_\mathrm{P}}$ for the \ac{PUAV}.

\subsubsection{Finding a guiding viewpoint}\label{sec:viewpoint}

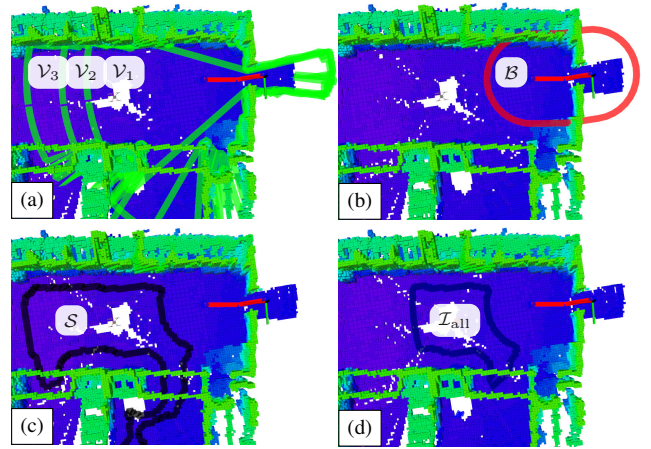
\begin{figure}[t]
  \centering
  \begin{tikzpicture}
    \node[anchor=north west,inner sep=0] (a) at (0cm, 0cm)
    {
      \includegraphics[width=0.5\linewidth, trim=2cm 8cm 2cm 6cm, clip=true]{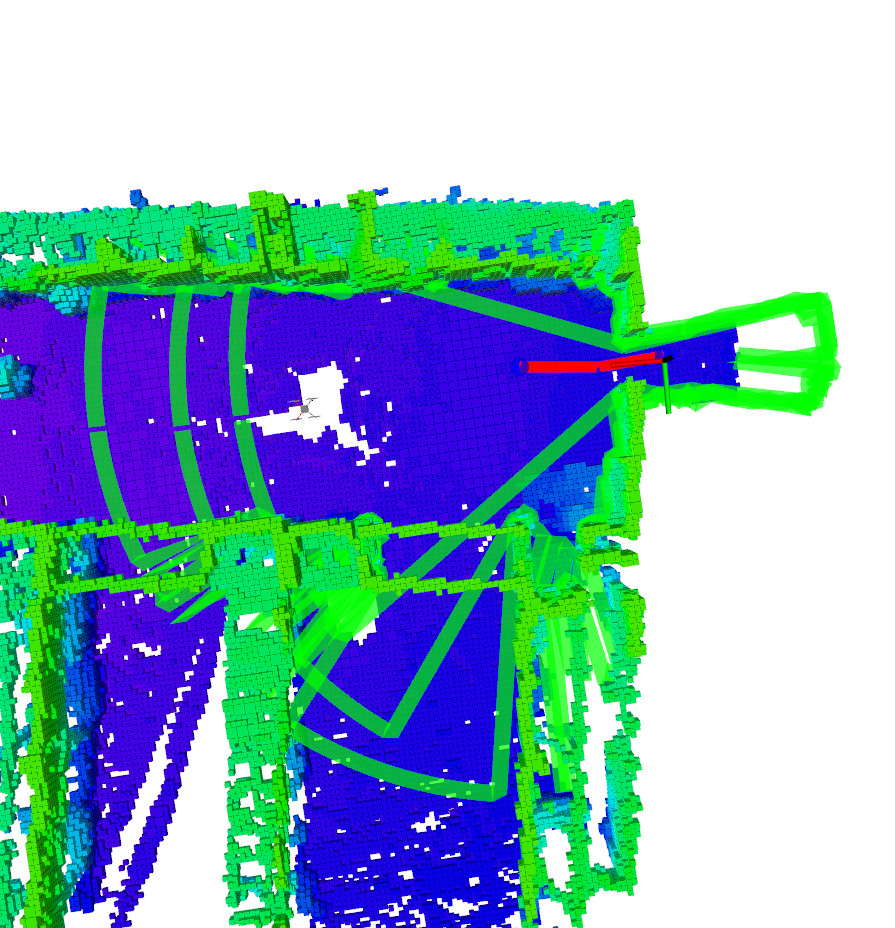}
    };
    \node[fill=white,draw=black,text=black, anchor=south west] at (a.south west) {\footnotesize (a)};

    \node[anchor=north west,inner sep=0] (b) at (4.35cm, 0cm)
    {
      \includegraphics[width=0.5\linewidth, trim=2cm 8cm 2cm 6cm, clip=true]{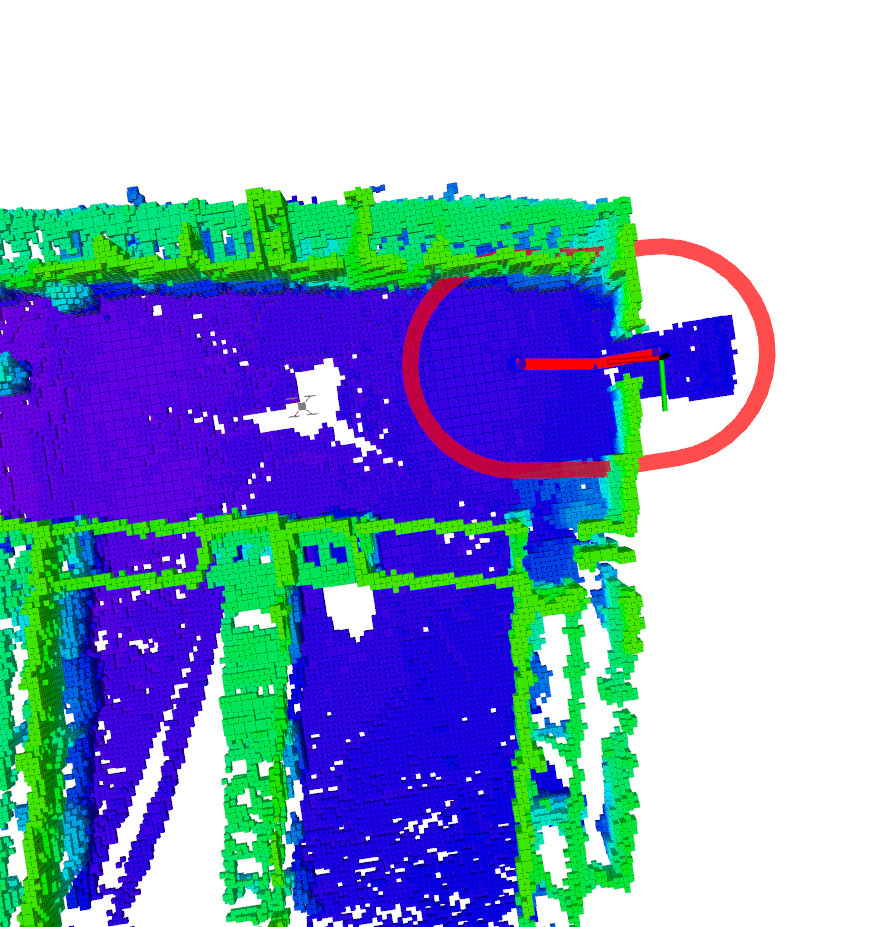}
    };
    \node[fill=white,draw=black,text=black, anchor=south west] at (b.south west) {\footnotesize (b)};

    \node[anchor=north west,inner sep=0] (d) at (0cm, -3.0cm)
    {
      \includegraphics[width=0.5\linewidth, trim=2cm 8cm 2cm 6cm, clip=true]{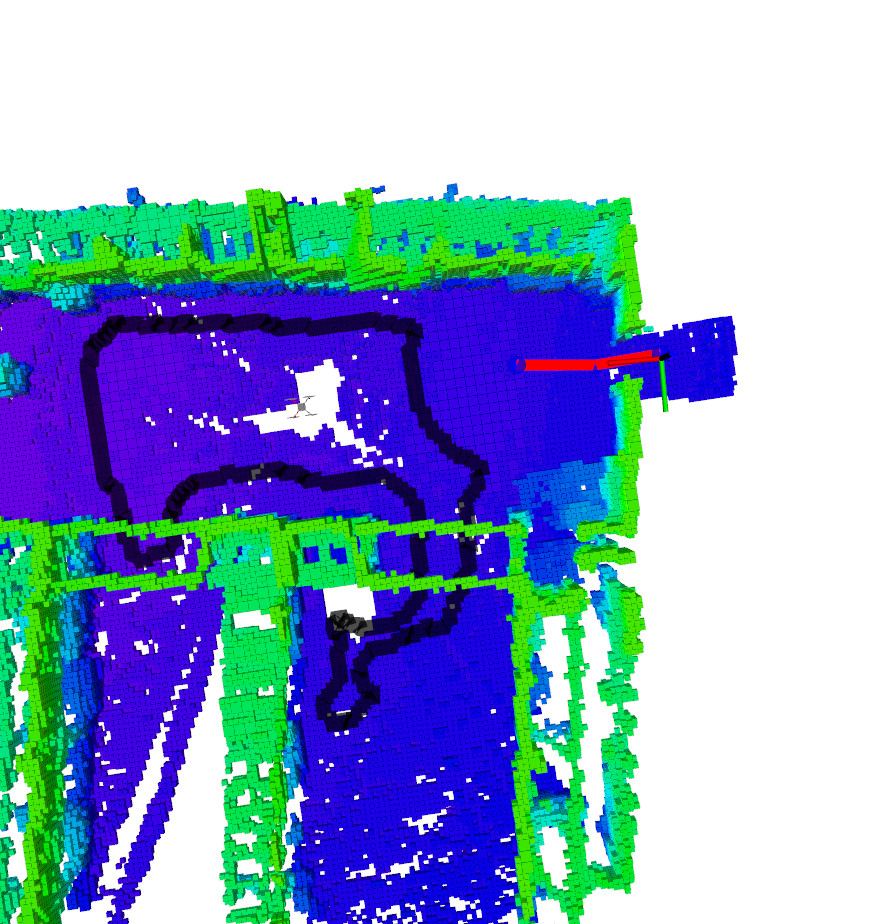}
    };
    \node[fill=white,draw=black,text=black, anchor=south west] at (d.south west) {\footnotesize (c)};

    \node[anchor=north west,inner sep=0] (e) at (4.35cm, -3.0cm)
    {
      \includegraphics[width=0.5\linewidth, trim=2cm 8cm 2cm 6cm, clip=true]{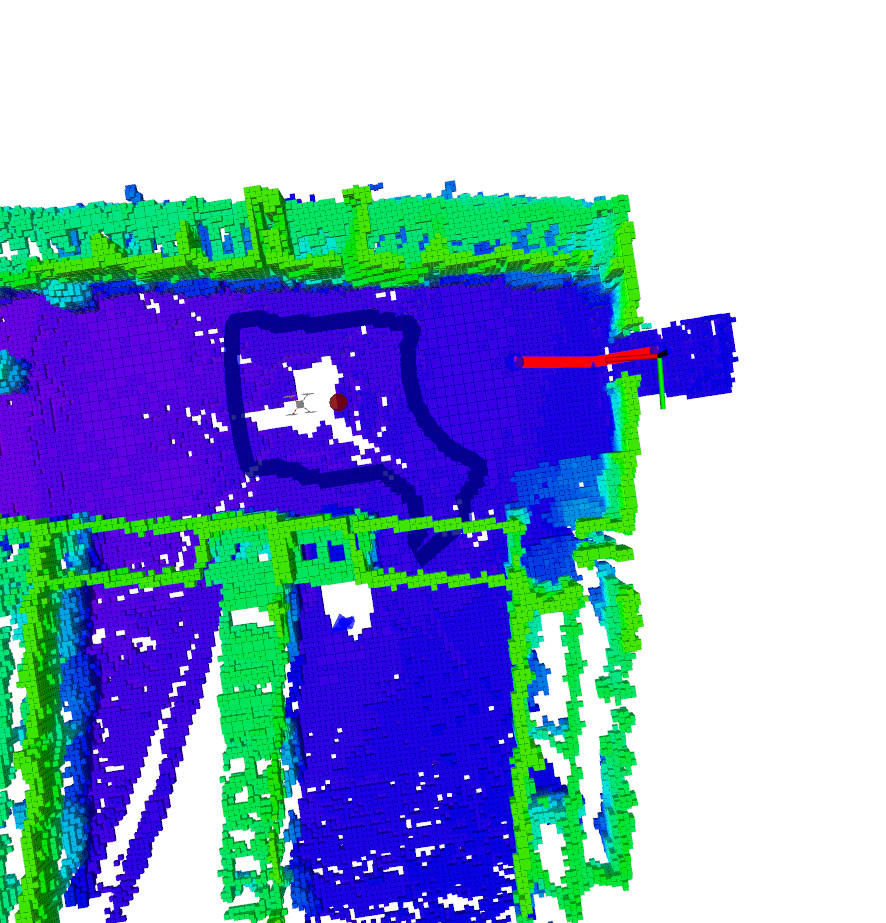}
    };
    \node[fill=white,draw=black,text=black, anchor=south west] at (e.south west) {\footnotesize (d)};

    \node[fill=white, rounded corners,opacity=0.9] (label1) at (1.5, -1.0) {
      \scriptsize
      \color{black}{$\mathcal{V}_1$}
    };
    \node[fill=white, rounded corners,opacity=0.9] (label1) at (1.0, -1.0) {
      \scriptsize
      \color{black}{$\mathcal{V}_2$}
    };
    \node[fill=white, rounded corners,opacity=0.9] (label1) at (0.5, -1.0) {
      \scriptsize
      \color{black}{$\mathcal{V}_3$}
    };

    \node[fill=white, rounded corners,opacity=0.9] (label1) at (6.65, -1.0) {
      \scriptsize
      \color{black}{$\mathcal{B}$}
    };

    \node[fill=white, rounded corners,opacity=0.9] (label1) at (0.8, -4.3) {
      \scriptsize
      \color{black}{$\mathcal{S}$}
    };

    \node[fill=white, rounded corners,opacity=0.9] (label1) at (5.9, -4.3) {
      \scriptsize
      \color{black}{$\mathcal{I}_\mathrm{all}$}
    };

  \end{tikzpicture}
  \caption{Example of guiding viewpoint selection. \ac{SUAV} path (red line) consists of 3 waypoints passing through the door in the upper right corner of the map. The algorithm constructs regions from which the waypoints are visible (a), a safety buffer around the \ac{SUAV} path (b), safe space w.r.t. obstacles for the \ac{PUAV} (c), and the final intersection $\mathcal{I}_\mathrm{all}$ (d).%
  }
  \label{fig:visib}
  \vspace{-1.3em}
\end{figure}

To maximize \ac{LOS} visibility throughout the guiding process, a viewpoint $\frames{L}{}{\vec{g}}_\mathrm{P}$ for the \ac{PUAV} is computed (see Alg.~\ref{alg:vis_planning} and Fig.~\ref{fig:visib}), such that it fulfills the following conditions:
\begin{enumerate}
  \item The \ac{PUAV} has \ac{LOS} visibility of as many \ac{SUAV} waypoints from the path beginning as possible. The areas with \ac{LOS} visibility of the specific \ac{SUAV} waypoints are represented as a set of non-convex polygons $\mathcal{V}_\mathrm{all}$.
  \item The closest obstacle distance is larger than a threshold $d_\mathrm{P}$.
    This creates a safe region, represented as a multi-polygon (collection of non-convex polygons with holes and non-intersecting boundaries) $\mathcal{S}$.
  \item The distance to the \ac{SUAV} path $\frames{L}{}{\mathcal{P}_\mathrm{S}}$ is larger than a desired safety threshold $d_\mathrm{buffer}$. The safety buffer around the \ac{SUAV} path is represented as a non-convex polygon $\mathcal{B}$.
\end{enumerate}
Alg.~\ref{alg:vis_planning} constructs 2D polygons representing the visibility regions $\mathcal{V}_\mathrm{all}$ by performing raycasting from each waypoint $\vec{p}_i$ to the surrounding voxels at the same altitude.
The algorithm calculates the difference of the visibility regions and the safety buffer $\mathcal{B}$, finds the closest safe region $\mathcal{S}_\mathrm{closest}$ within $\bigcup \mathcal{V}_\mathrm{all}$, and looks for the intersection of $\mathcal{S}_\mathrm{closest}$ with as many consecutive visibility regions $\mathcal{V}_{0\dots k}$ as possible.
If such an intersection exists, the viewpoint $\frames{L}{}{\vec{g}}_\mathrm{P}$ is selected as the closest one from the interior points furthest from the boundaries, i.e., the poles of inaccessibility (function POI() in Alg.~\ref{alg:vis_planning}), of the obtained intersection (the intersection may be a collection of polygons).
The polygonal operations were implemented using the \textit{boost::geometry} C++ library\footnote{\url{https://www.boost.org/libs/geometry/}}.

\subsubsection{Guiding}

The \textit{Multi-UAV path planner} transforms the \ac{SUAV} path $\frames{L}{}{\mathcal{P}_\mathrm{S}}$ to the \ac{SUAV} body frame ${S}$ and removes all points $\frames{S}{}{\vec{p}_i} \in \frames{S}{}{\mathcal{P}_\mathrm{S}}$ such that $\left|\left|\frames{S}{}{\vec{p}_i}\right|\right|_2 < \delta$, where $\delta$ is a distance threshold for considering the path waypoint as visited.
The remaining path is transmitted to the \ac{SUAV} over the wireless network and saved for the next guiding step.
It is worth mentioning that the planned path could instead be transmitted in the original frame ${L}$ along with the transformation $\frames{V}{L}{\mat{T}}$ and transformed on board the \ac{SUAV}.
Transforming the path on board the \ac{PUAV} minimizes the requirements imposed on the \ac{SUAV}, making the approach compatible with a wide range of \acp{UAV}.
The only requirements on the \ac{SUAV} are the capability to transmit its local odometry in the local frame ${V}$ and follow paths specified relatively to its current pose.

\subsection{Relative pose estimation}
\vspace{-0.1em}
Accurate relative pose estimation between the \acp{UAV} is crucial for reliable performance of the guiding approach, and uncertainty in the relative pose represents one of the \bl{possible} sources of guiding errors.
The estimation approach needs to accurately provide the transformation $\frames{S}{L}{\mat{T}}$ required for transforming the reference path from the local frame ${L}$ to the \ac{SUAV} body frame ${S}$.
\bl{In the proposed approach, we apply relative pose estimation based on the fusion of \ac{lidar} detections with \ac{VIO} data received over the wireless network.
The approach for \ac{UAV} detection from \ac{lidar} data is described in~\cite{vrbaOnboardLiDARbasedFlying2023} in detail.
The \ac{lidar} detections provide precise 3D positions of the \ac{SUAV}, and the fusion with \ac{VIO} data provides orientation of the \ac{SUAV} and keeps track of the \ac{SUAV} when the detections are lost.
The fusion approach is described in~\cite{pritzlFusionVisualInertialOdometry2023} in detail and does not require any apriori knowledge about the \ac{UAV} positions nor orientations.
The relative transformation $\frames{V}{L}{\mat{T}}$ is obtained by solving a \ac{NLS} problem aligning corresponding \ac{UAV} trajectories observed in a sliding window of the \ac{lidar} detections and \ac{VIO} data.
The periodically-updated transformation $\frames{V}{L}{\mat{T}}$, \ac{lidar}-based detections, and \ac{VIO} data are utilized in a Kalman-filter-based approach to estimate the transformation $\frames{S}{L}{\mat{T}}$.}
\bl{The \ac{UAV} detection approach~\cite{vrbaOnboardLiDARbasedFlying2023} was improved by utilizing reflective markers on the legs of the \ac{SUAV}.}
The employed 3D \ac{lidar} sensor provides the reflectivity of each measured data point, hence a thresholding-based filter was applied to filter out possible false detections, and thus increase the reliability of \ac{UAV} detections in cluttered environments.



\section{EXPERIMENTAL VERIFICATION}
\vspace{-0.1em}
\label{sec:experiments}

Two different quadrotor platforms (see Fig.~\ref{fig:motivation}) were utilized in the experiments.
The \ac{PUAV} was built upon the Holybro X500 frame and was \SI{0.7}{m} $\times$ \SI{0.7}{m} wide including propellers.
The \ac{SUAV} was built upon the DJI F330 frame and was \SI{0.45}{m} $\times$ \SI{0.45}{m} wide including propellers.
Both \acp{UAV} carried the Intel NUC 10i7FNH onboard computer with the Intel Core i7 10710U CPU, 16 GB of RAM, and a Wi-Fi module.
Both \acp{UAV} used the Pixhawk 4 flight controller.
The \ac{PUAV} was equipped with the Ouster OS0-128 Rev C 3D \ac{lidar}.
The \ac{lidar} has a 360$^\circ$ horizontal and 90$^\circ$ vertical \ac{FOV} and produces scans with a resolution of 1024 $\times$ 128 beams at a rate of 10 Hz.
For mapping, the \ac{lidar} scans were downsampled by a factor of 4 to ensure the real-time performance of the entire software stack on board the \acp{UAV}.
Detailed information about the \ac{UAV} hardware is available in~\cite{MRS2022ICUASHW, HertJINTHWpaper}.

The \ac{PUAV} utilized the LOAM algorithm~\cite{zhangLOAMLidarOdometry2014} for self-localization.
The \ac{SUAV} carried a front-facing RealSense T265 tracking camera and utilized its stereo fisheye image and \ac{IMU} output for \ac{VIO} using the OpenVINS algorithm~\cite{Geneva2020ICRA}.
The \ac{SUAV} was equipped with reflective markers on its legs to aid the \ac{lidar}-based relative localization.
The software on board both \acp{UAV} was based on Ubuntu 20.04, \ac{ROS} 1, and the MRS UAV system~\cite{bacaMRSUAVSystem2021}.
Both \acp{UAV} were connected to a single Wi-Fi access point for mutual communication, and the Nimbro network was used for transporting the \ac{ROS} topics over the wireless network.
The system time of the \acp{UAV} was synchronized using the \textit{chrony} implementation of the \ac{NTP}.

\bl{
\subsection{Simulated evaluation of guiding accuracy}
\vspace{-0.1em}

\begin{figure}[t]
  \centering
  \begin{tikzpicture}
    \node[anchor=north west,inner sep=0] (a) at (0, 0)
    {
      \includegraphics[width=0.6\linewidth, trim=10cm 5cm 18cm 12cm, clip=true]{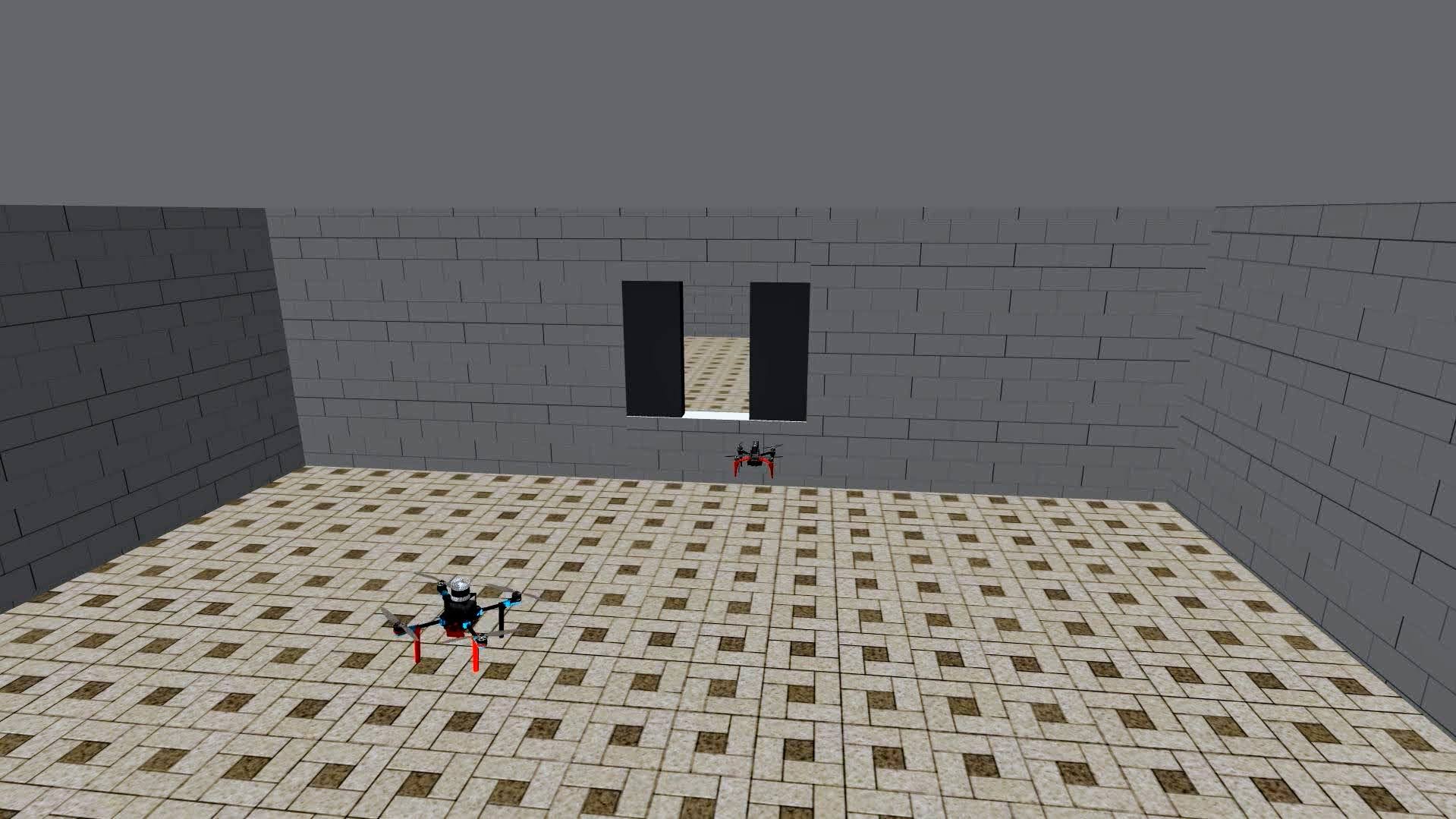}
    };

    \node[anchor=north west,inner sep=0] (a) at (5.1cm, 0)
    {
      \includegraphics[width=0.4\linewidth, trim=0.2cm 0.35cm 0.3cm 0.3cm, clip=true]{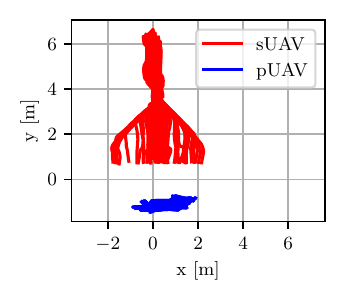}
    };

    \node[fill=white, rounded corners,opacity=0.9] (label1) at (2.4, -2.2) {
      \scriptsize
      \color{black}{pUAV}
    };

    \node[fill=white, rounded corners,opacity=0.9] (label2) at (2.5, -1.4) {
      \scriptsize
      \color{black}{sUAV}
    };

  \end{tikzpicture}
  \caption{\bl{The simulation with a gap of varying width and \ac{UAV} trajectories from the test of the proposed guiding approach.}%
  }
  \label{fig:gazebo}
  \vspace{-1.3em}
\end{figure}





\begin{figure}[t]
  \centering
  \begin{tikzpicture}
    \node[anchor=north west,inner sep=0] (a) at (0, 0)
    {
      \includegraphics[width=1.0\linewidth, trim=0.51cm 0.29cm 1.1cm 0.7cm, clip=true]{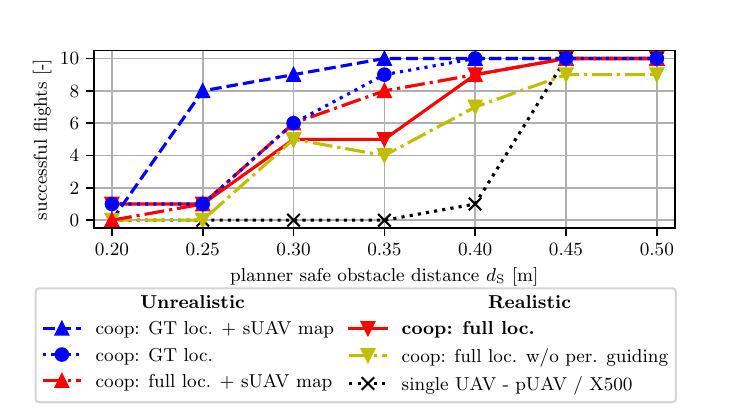}
    };

  \end{tikzpicture}
  \caption{The number of successful flights through the gap depending on the planner's distance $d_\mathrm{S}$ for each simulated configuration.
  The legend is ordered based on the total number of successful flights.
  \textit{Single \ac{UAV}} represents a \ac{UAV} carrying a 3D LiDAR.
  \textit{Coop} represents guiding \ac{SUAV} by \ac{PUAV}.
  \textit{GT loc.} means ground-truth self-localization and ground-truth relative localization.
  \textit{Full loc.} denotes localization from fully simulated onboard sensors.
  \textit{\ac{SUAV} map} means guiding the \ac{SUAV} by \ac{PUAV} using a map constructed from a 3D \ac{lidar} on board the \ac{SUAV} (in reality too heavy to be carried by the \ac{SUAV}).
  \textit{w/o per. guiding} means sending the path to the \ac{SUAV} only once, without periodic guiding.
  }
  \label{fig:simulation_successes}
  \vspace{-1.3em}
\end{figure}

The main factors influencing the guiding accuracy were identified in a simulation study using a two-room environment connected by a gap with varying width built in the Gazebo simulator (see Fig.~\ref{fig:gazebo}).
The gap width was set to
\begin{equation}w_\mathrm{G} = 2\times d_\mathrm{S} + 1.5\times\mathcal{M}.resolution,\end{equation}
  where $d_\mathrm{S}$ is the planner's safe distance for \ac{SUAV} and $\mathcal{M}.resolution = \SI{0.1}{m}$ is the voxel size of the map.
Multiple configurations of the framework were compared.
For each configuration, the ability to fly through the gap was tested for a set of planner's safe distances $d_\mathrm{S}$.
For each distance $d_\mathrm{S}$, 10 runs of simulation with randomized initial \ac{UAV} positions were done, randomizing the direction of the necessary \ac{UAV} movement towards the gap and the alignment of the map grid with the environment.
A simulation run was considered successful if the \ac{SUAV} flew along planned paths through the gap and back to predefined goal position without colliding with the obstacles.

Fig.~\ref{fig:simulation_successes} compares the number of successful flights per safe distance $d_\mathrm{S}$ among the tested configurations.
The effect of localization errors was compared on guiding with ground-truth localization and full localization using LOAM~\cite{zhangLOAMLidarOdometry2014}, OpenVINS~\cite{Geneva2020ICRA}, and \ac{lidar} detections~\cite{pritzlFusionVisualInertialOdometry2023}, identical to the real-world setup.
The effect of mapping errors was compared on guiding using a map constructed on board \ac{SUAV} (from onboard 3D \ac{lidar} too heavy to be carried in reality), and a map constructed on board the \ac{PUAV}.
The configuration \textit{coop: full loc.} in Fig.~\ref{fig:simulation_successes} corresponds to the real-world setup.

The guiding accuracy declines with increasing inaccuracy of mapping caused by the increase of the gap between adjacent \ac{lidar} rays (\textit{GT loc.} vs \textit{GT loc. + sUAV map} in Fig.~\ref{fig:simulation_successes}).
The mutual \ac{UAV} distance was approximately \SI{5}{m} when the \ac{SUAV} was passing through the gap.
At the distance of \SI{5}{m}, the gap between adjacent \ac{lidar} rays in the horizontal plane is \SI{0.12}{m}, assuming the 256 horizontal samples of the downsampled \ac{lidar} scan.
The \ac{lidar}-based relative localization accuracy also depends on the mutual \ac{UAV} distance.
The \ac{MAE} of \ac{SUAV} localization based on the \ac{lidar} detections within \SI{1}{m} from the center of the gap was \SI{0.10}{m}, and the mean error in the axis parallel to the gap was \SI{0.054}{m}.
By comparing the \textit{coop: GT loc. + sUAV map} configuration and the \textit{coop: full loc.} configuration, we see that the distance $d_\mathrm{S}$ for which the \ac{SUAV} was able to achieve 90\% success rate increased by \SI{0.1}{m} due to these mapping and localization errors.

The \textit{coop: full loc. w/o per. guiding} configuration shows that sending the path to the \ac{SUAV} only once without periodic guiding results in decreased performance due to localization errors.
The \textit{single UAV} configuration shows the gap width necessary for the \ac{PUAV} to pass through.
The simulations have proven the viability of guiding the \ac{SUAV} by the \ac{PUAV}, showing that the \ac{SUAV} can be guided through much smaller gaps than the \ac{PUAV} itself would be able to pass through.

}

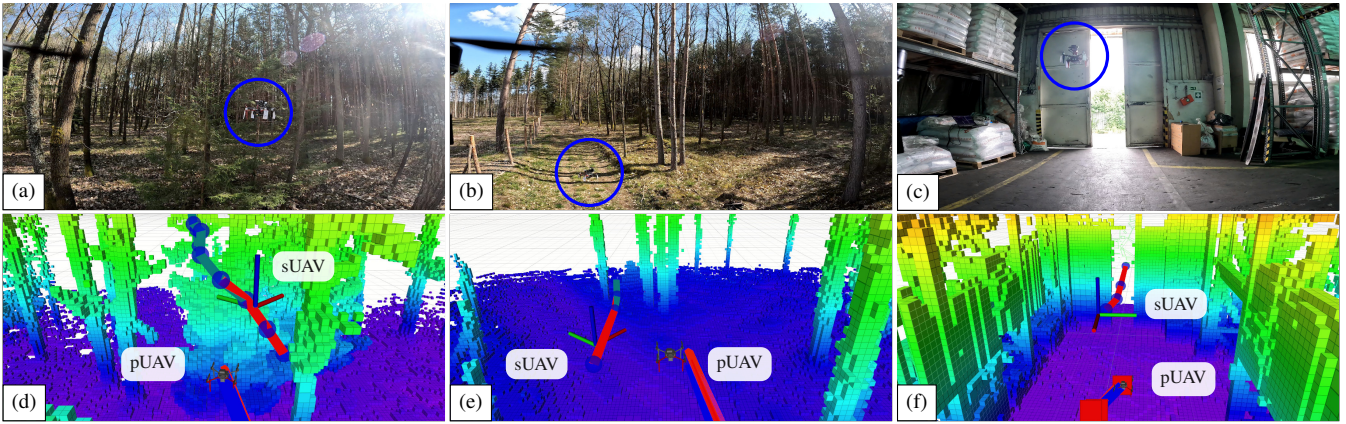
\begin{figure*}[t]
  \centering
  \begin{tikzpicture}
    \node[anchor=north west,inner sep=0,draw=black] (a) at (0, 0)
    {
      \includegraphics[width=0.33\linewidth, trim=0cm 1.7cm 0cm 1.7cm, clip=true]{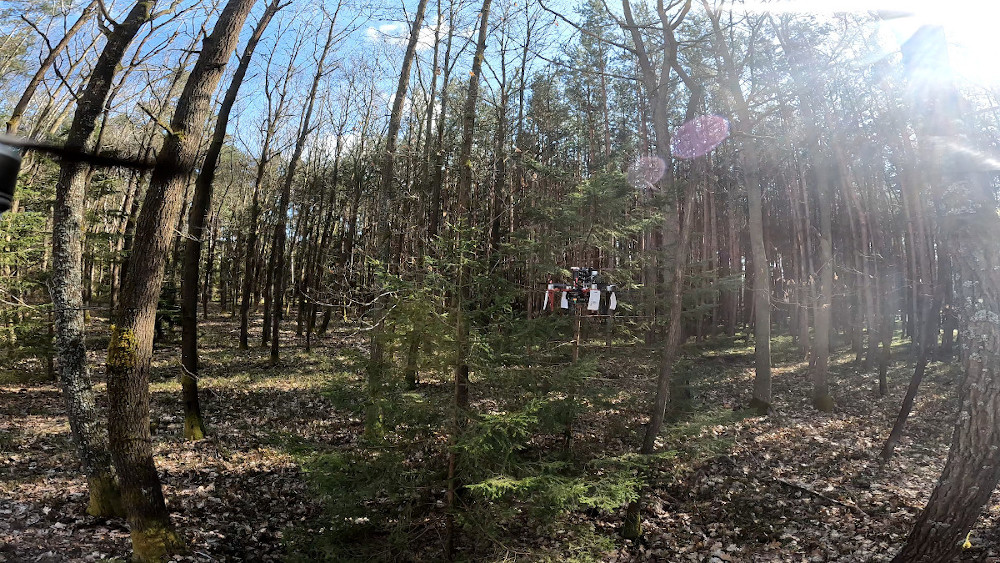}
    };
    \node[fill=white,draw=black,text=black, anchor=south west] at (a.south west) {\footnotesize (a)};

    \node[anchor=north west,inner sep=0,draw=black] (b) at (5.945cm, 0cm)
    {
      \includegraphics[width=0.33\linewidth, trim=0cm 1.7cm 0cm 1.7cm, clip=true]{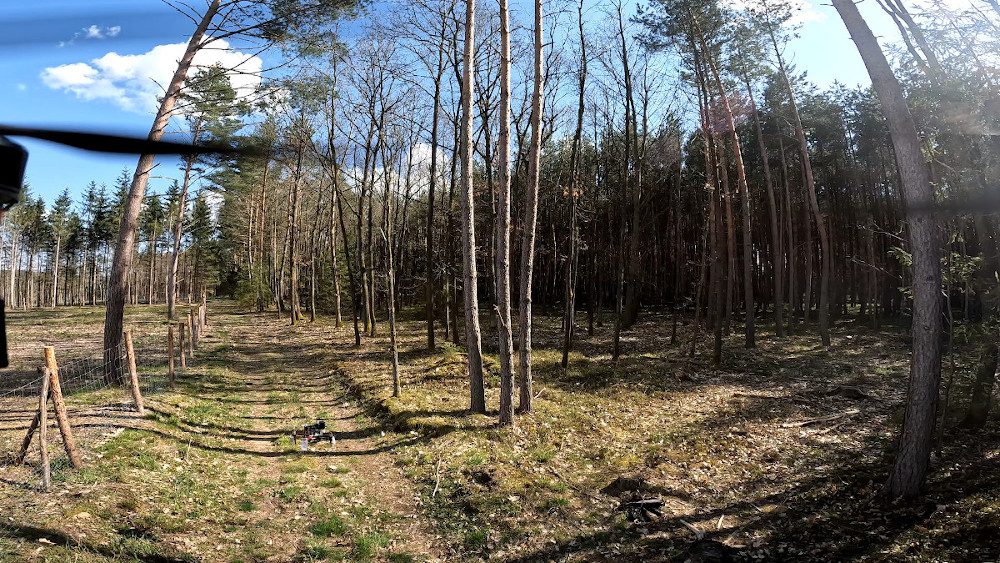}
    };
    \node[fill=white,draw=black,text=black, anchor=south west] at (b.south west) {\footnotesize (b)};

    \node[anchor=north west,inner sep=0,draw=black] (c) at (11.89cm, 0)
    {
      \includegraphics[width=0.33\linewidth, trim=0cm 2cm 0cm 2cm, clip=true]{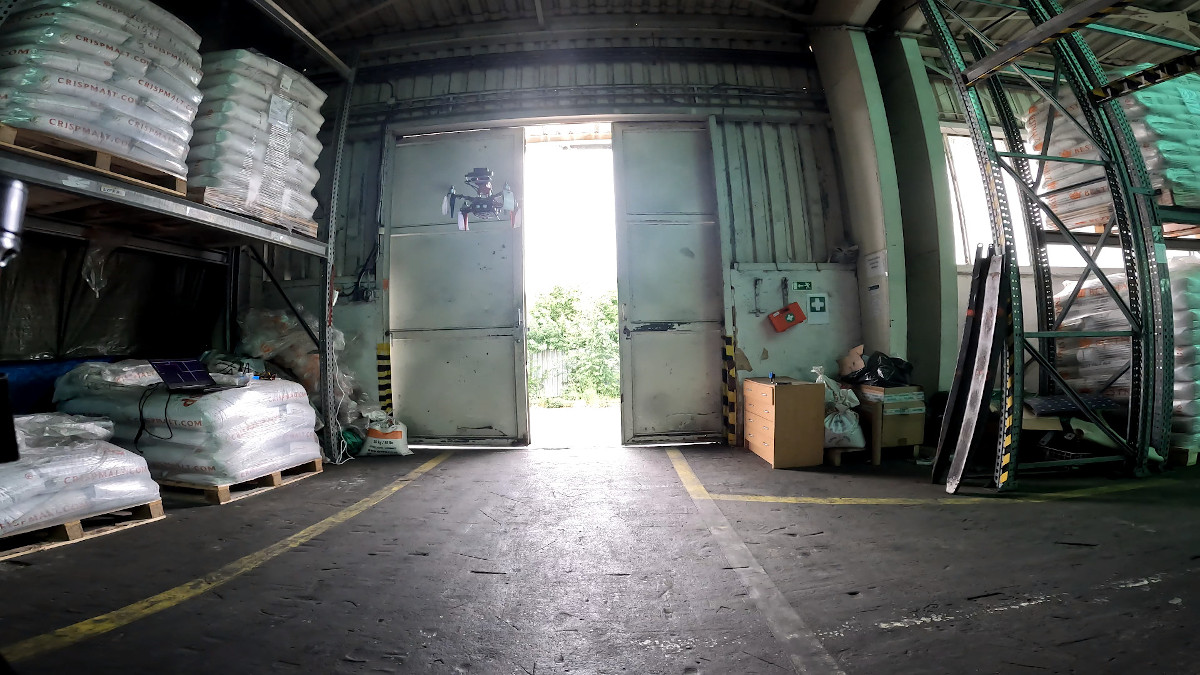}
    };
    \node[fill=white,draw=black,text=black, anchor=south west] at (c.south west) {\footnotesize (c)};

    \node[anchor=north west,inner sep=0,draw=black] (d) at (0cm, -2.8cm)
    {
      \includegraphics[width=0.33\linewidth, trim=0cm 4cm 0cm 0.0cm, clip=true]{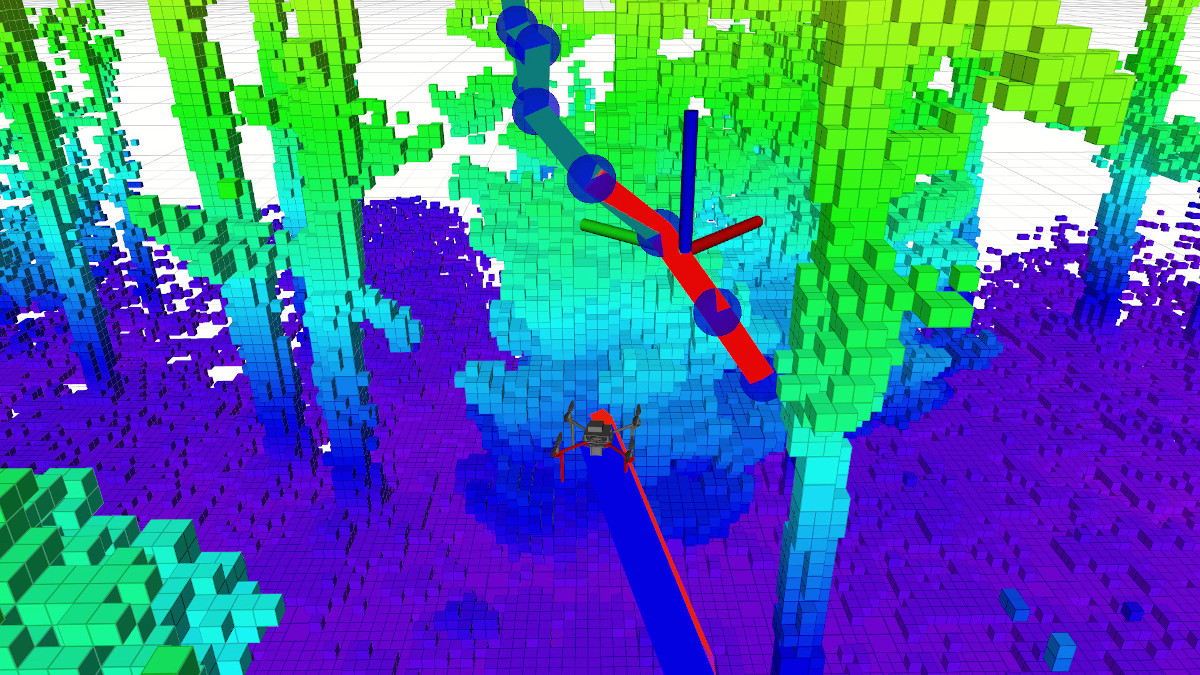}
    };
    \node[fill=white,draw=black,text=black, anchor=south west] at (d.south west) {\footnotesize (d)};

    \node[anchor=north west,inner sep=0,draw=black] (e) at (5.945cm, -2.8cm)
    {
      \includegraphics[width=0.33\linewidth, trim=0cm 2cm 0cm 2cm, clip=true]{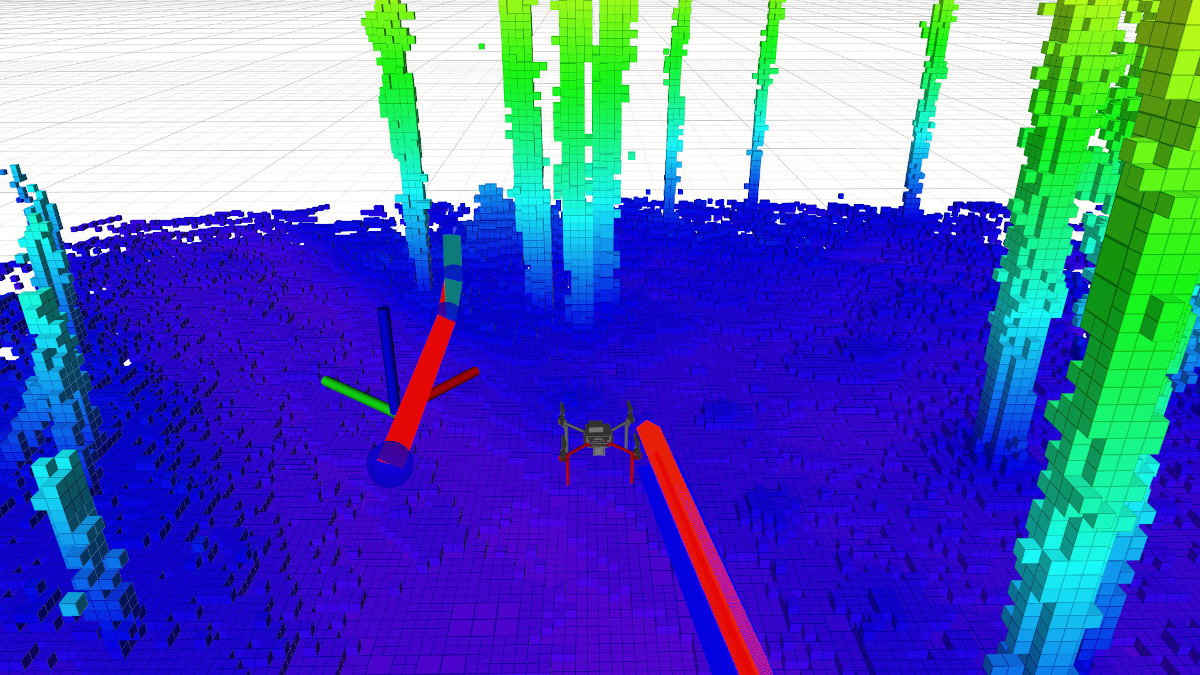}
    };
    \node[fill=white,draw=black,text=black, anchor=south west] at (e.south west) {\footnotesize (e)};

    \node[anchor=north west,inner sep=0,draw=black] (f) at (11.89cm, -2.8cm)
    {
      \includegraphics[width=0.33\linewidth, trim=0cm 1cm 0cm 3cm, clip=true]{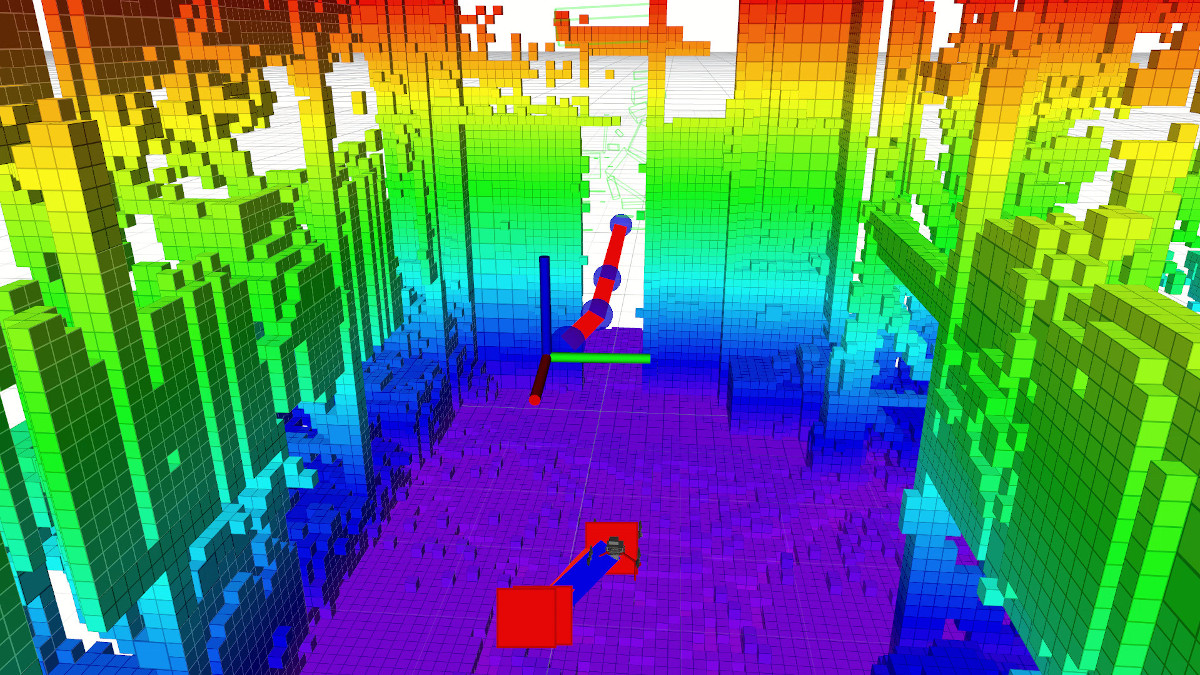}
    };
    \node[fill=white,draw=black,text=black, anchor=south west] at (f.south west) {\footnotesize (f)};

    \node[highlight] (sec_a) at (3.4,-1.45) {};
    \node[highlight] (sec_b) at (7.8,-2.25) {};
    \node[highlight] (sec_c) at (14.25,-0.7) {};

    \node[fill=white, rounded corners,opacity=0.9] (label1) at (2.0, -4.8) {
      \scriptsize
      \color{black}{pUAV}
    };
    \node[fill=white, rounded corners,opacity=0.9] (label2) at (4.0, -3.5) {
      \scriptsize
      \color{black}{sUAV}
    };

    \node[fill=white, rounded corners,opacity=0.9] (label3) at (9.8, -4.8) {
      \scriptsize
      \color{black}{pUAV}
    };
    \node[fill=white, rounded corners,opacity=0.9] (label4) at (7.1, -4.8) {
      \scriptsize
      \color{black}{sUAV}
    };

    \node[fill=white, rounded corners,opacity=0.9] (label5) at (15.7, -5.0) {
      \scriptsize
      \color{black}{pUAV}
    };
    \node[fill=white, rounded corners,opacity=0.9] (label6) at (15.6, -4.0) {
      \scriptsize
      \color{black}{sUAV}
    };

  \end{tikzpicture}
  \caption{Real-world experiments: images from the \ac{PUAV} onboard camera (a-c) and visualization of the occupancy map and planned paths at the corresponding moments from similar views behind the \ac{PUAV} (d-f). The \ac{SUAV} is marked by blue circles.%
  }
  \label{fig:experiments}
  \vspace{-1.3em}
\end{figure*}

\subsection{Real-world cooperative flight through a forest}\label{sec:forest}
\vspace{-0.1em}

\begin{figure}[t]
  \centering
  \begin{tikzpicture}
    \node[anchor=north west,inner sep=0] (a) at (0, 0)
    {
      \includegraphics[width=1.0\linewidth, trim=0cm 0cm 0cm 0cm, clip=true]{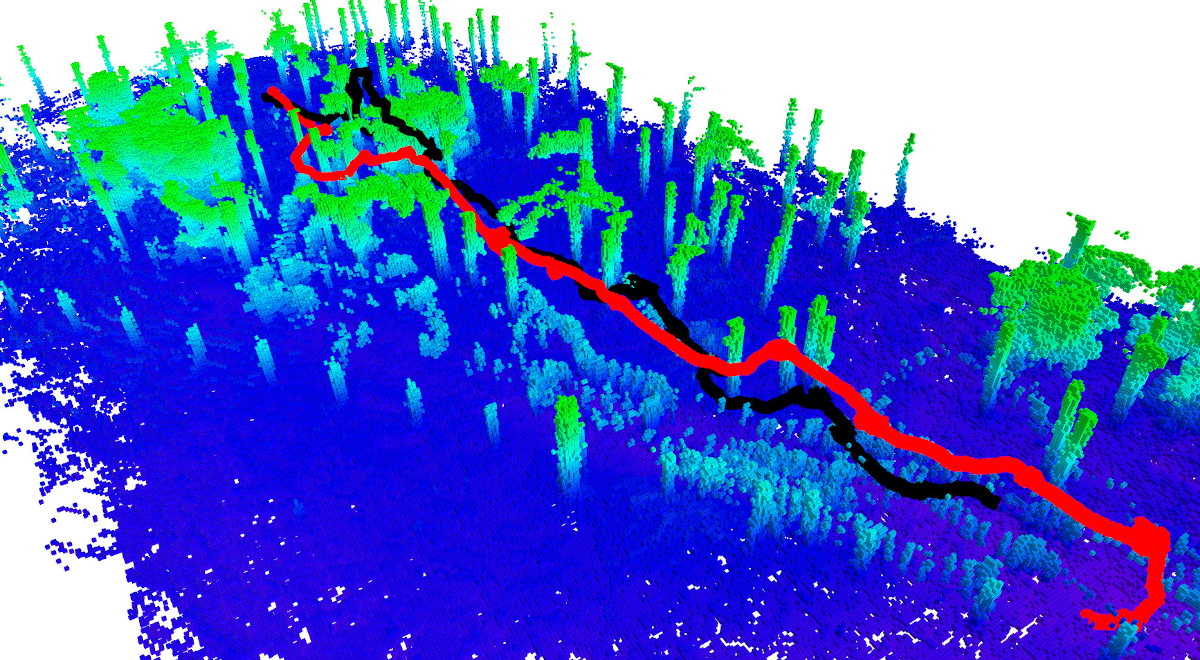}
    };

    \node[fill=white, rounded corners,opacity=0.9] (label1) at (7.0, -4.4) {
      \scriptsize
      \color{black}{pUAV start}
    };

    \node[fill=white, rounded corners,opacity=0.9] (label2) at (6.6, -3.9) {
      \scriptsize
      \color{black}{sUAV start}
    };

  \end{tikzpicture}
  \caption{Occupancy map from the forest flight and the \ac{UAV} trajectories (\ac{PUAV} - red, \ac{SUAV} - black). For clarity, the map was sliced at the maximum height of \SI{4}{m} above the ground.%
  }
  \label{fig:forest_octomap}
  \vspace{-0.5em}
\end{figure}

In the next set of experiments, the \ac{PUAV} was tasked to guide the \ac{SUAV} through a forest (see Fig.~\ref{fig:experiments} and the video attachment).
The guiding process was triggered repeatedly, and the \ac{SUAV} goal pose was generated by incrementing the \ac{SUAV} pose by \SI{4}{m} in the direction of the \textit{x}-axis each time the guiding process finished (switched to the \textit{GOAL REACHED} state).
Table~\ref{tab:forest_params} contains parameters of the Algorithms \ref{alg:planning} and \ref{alg:vis_planning} utilized in the forest flights.
Five separate flights were performed.
Table~\ref{tab:forest_distance} contains the distances traveled during these flights.
The \acp{UAV} performed all of these flights successfully, reaching the border of the allowed operation space.
Fig.~\ref{fig:forest_octomap} shows the occupancy map built during one of the flights along with the trajectories traversed by the two \acp{UAV}.

\begin{table}[t]
\scriptsize
\begin{center}
  \caption{Parameters from the forest flight experiments.}
  \begin{tabularx}{0.95\linewidth}{l *{3}{Y}} 
\toprule 
    Parameter & Symbol & Value \\ \midrule
    voxel size & $\mathcal{M}.resolution$ & \SI{0.1}{m} \\
    \ac{PUAV} safe distance & $d_\mathrm{P}$ & \SI{0.9}{m} \\
    \ac{SUAV} safe distance & $d_\mathrm{S}$ & \SI{0.8}{m} \\
    \ac{PUAV} occ. space width & $w_\mathrm{P}$ & \SI{1.5}{m} \\
    \ac{PUAV} occ. space height & $h_\mathrm{P}$ & \SI{10}{m} \\
    \ac{SUAV} occ. space width & $w_\mathrm{S}$ & \SI{1.3}{m} \\
    \ac{SUAV} occ. space height & $h_\mathrm{S}$ & \SI{10}{m} \\
    raycasting sample count & $n_\mathrm{samples}$ & 500 \\
    raycasting max length & $d_\mathrm{ray}$ & \SI{6}{m} \\
    min. safe distance from \ac{SUAV} path & $d_\mathrm{buffer}$ & \SI{2}{m} \\

    \bottomrule
\end{tabularx}
  \label{tab:forest_params}
\end{center}
  \vspace{-1.5em}
\end{table}

\begin{table}[t]
\scriptsize
\begin{center}
  \caption{2D straight-line distances traveled from the start of the guiding process by the \ac{SUAV} in the forest. The distances were calculated from the relative localization data.}
  \begin{tabularx}{0.95\linewidth}{l *{6}{Y}} 
\toprule 
    Flight number & 1 & 2 & 3 & 4 & 5 \\ \midrule
    Distance traveled [m] & 31.6 & 36.5 & 35.1 & 41.6 & 35.0 \\
    \bottomrule
\end{tabularx}
  \label{tab:forest_distance}
\end{center}
  \vspace{-1.5em}
\end{table}

The median processing time of the entire Alg.~\ref{alg:planning} was \SI{8.90}{s}.
The median processing times of the individual steps were \SI{0.89}{s} for the guiding viewpoint selection (Alg.~\ref{alg:vis_planning}, Alg.~\ref{alg:planning} line 9), \SI{0.60}{s} for \ac{PUAV} path planning (Alg.~\ref{alg:planning} line 17), \SI{5.33}{s} for \ac{SUAV} path planning (Alg.~\ref{alg:planning} line 5), and the rest of the time was spent copying the occupancy maps and inserting occupied space into them.
The processing time of \ac{PUAV} path planning was much lower since the \ac{PUAV} goal was always located in a safe space.
In contrast, the \ac{SUAV} goal was calculated by incrementing the \ac{SUAV} pose and thus often laid too close to an obstacle, causing the planner to search for a path until a timeout threshold was reached.
In a practical application, a feasible goal can be pre-selected, or the planning timeout can be decreased to reduce the planning time.
For safety reasons, the planning process was performed only while the \acp{UAV} were hovering, but it can be easily triggered while the \ac{SUAV} is still moving.


\subsection{Real-world guiding through a narrow gap}
\vspace{-0.1em}

The last set of experiments, performed in an industrial warehouse environment, aims to evaluate the performance of the proposed framework in navigation of the \ac{SUAV} through extremely narrow openings.
The \ac{PUAV} was tasked to guide the \ac{SUAV} through a narrow gap whose width was gradually decreased together with the safety distance of the path planner.
The algorithms utilized the same parameters as in Table~\ref{tab:forest_params}, except for the \ac{SUAV} safe distance $d_\mathrm{S}$, which was decreased according to the width of the gap.
Both \acp{UAV} started flying inside the warehouse, and the \ac{SUAV} was guided through the narrow gap out of the building and back multiple times.
Table~\ref{tab:table_rudna} shows the results of this experiment.
The proposed guiding approach reached its limit when passing through a \SI{0.9}{m} wide gap when the \ac{SUAV} collided with the door during the fifth attempted pass due to an inaccuracy in the relative localization.
\bl{The results correspond to the performance observed in the simulations (see Fig.~\ref{fig:simulation_successes}).
}

\begin{table}[t]
\scriptsize
\begin{center}
  \caption{Results of guiding the \ac{SUAV} through narrow gaps with varying width. A single pass represents flight through the gap either out of the building or into the building.}
  \begin{tabularx}{0.95\linewidth}{l *{5}{Y}} 
\toprule 
    Gap width [m] & 1.2 & 1.1 & 1.0 & 0.9 \\
    \ac{SUAV} safe distance $d_\mathrm{S}$ [m] & 0.5 & 0.45 & 0.4 & 0.4 \\ \midrule
    Successful / attempted passes [-] & 2/2 & 2/2 & 6/6 & 4/5 \\
    \bottomrule
\end{tabularx}
  \label{tab:table_rudna}
\end{center}
  \vspace{-1.5em}
\end{table}

\begin{figure}[t]
  \centering
  \begin{tikzpicture}
    \node[anchor=north west,inner sep=0] (a) at (0, 0)
    {
      \includegraphics[width=1.0\linewidth, trim=0.5cm 0.2cm 0.3cm 0.81cm, clip=true]{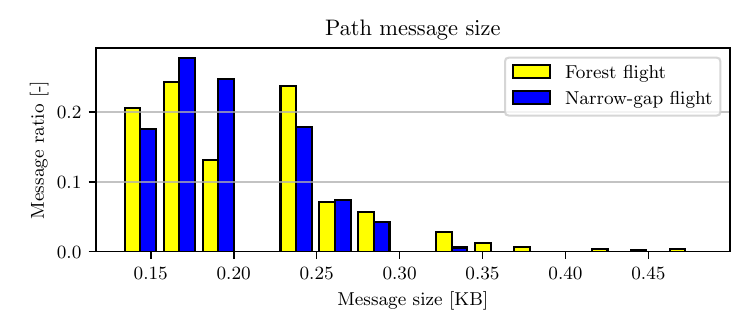}
    };


  \end{tikzpicture}
  \caption{\bl{Path message sizes during the real-world experiments.}%
  }
  \label{fig:message_sizes}
  \vspace{-1.3em}
\end{figure}

\bl{\subsection{Communication analysis}
\vspace{-0.1em}

Table~\ref{tab:table_communication} analyzes the wireless communication bandwidth required by the guiding approach.
The table shows message sizes recorded during the real-world experiments and the required maximal and minimal bandwidth calculated based on the transmission rates.
The data transmitted from the \ac{SUAV} to \ac{PUAV} consisted of the \ac{VIO} output $\frames{V}{S}{\mat{T}}$.
The data sent from the \ac{SUAV} to \ac{PUAV} were the desired paths $\frames{S}{}{\mathcal{P}_{\mathrm{S}}}$.

The sizes of the path messages depend on the number of contained points.
Fig.~\ref{fig:message_sizes} shows histograms of the message sizes from real-world experiments.
The communication bandwidth utilized by the guiding approach is minimal and leaves enough capacity for application-specific sensory data or commands for guiding multiple \acp{SUAV}.
}

\begin{table}[t]
\scriptsize
\begin{center}
  \caption{\bl{Analysis of the wireless communication requirements during the real-world experiments.}}
  \begin{tabularx}{0.95\linewidth}{l *{4}{Y}} 
\toprule 
    & & \ac{SUAV}$\rightarrow$\ac{PUAV} & \ac{PUAV}$\rightarrow$\ac{SUAV}  \\
    & & odometry & path  \\ \midrule
    \tabularxmulticolumncentered{2}{Y}{Rate [Hz]} & 2 & 5 \\
    \multirow{2}{*}{Message size [KB]} & min & 0.725 & 0.132 \\
     & max & 0.725 & 0.484 \\
    \multirow{2}{*}{Bandwidth [KBps]} & min & 1.45 & 0.66 \\
     & max & 1.45 & 2.42 \\
    \bottomrule
\end{tabularx}
  \label{tab:table_communication}
\end{center}
  \vspace{-1.5em}
\end{table}

\section{CONCLUSIONS}
\vspace{-0.1em}

A novel cooperative guiding approach enabling offloading the necessity to carry accurate but large and heavy sensors from a miniature \ac{UAV} to another member of a \ac{UAV} team while preserving the desired obstacle avoidance capabilities and reliable localization was proposed in this paper.
To achieve such tight cooperation, the more-capable primary \ac{UAV} performs occupancy mapping of the environment, relative localization of the less-capable secondary \ac{UAV}, planning of collision-free paths for both the secondary \ac{UAV} and itself, and guidance of the secondary \ac{UAV} through the environment.
To achieve reliable cooperative navigation in narrow passages, a reliable relative-localization-aware planner was proposed and integrated into the complex system.
The feasibility of the proposed approach was demonstrated in multiple real-world experiments in unknown cluttered \ac{GNSS}-denied environments with all the algorithms running on board the \acp{UAV} with no external localization system nor external computational resources utilized.

\label{sec:conclusion}









\begin{acronym}
  \acro{GPS}[GPS]{Global Positioning System}
  \acro{CNN}[CNN]{Convolutional Neural Network}
  \acro{MAV}[MAV]{Micro Aerial Vehicle}
  \acro{UAV}[UAV]{Unmanned Aerial Vehicle}
  \acro{UGV}[UGV]{Unmanned Ground Vehicle}
  \acro{UV}[UV]{ultraviolet}
  \acro{UVDAR}[\emph{UVDAR}]{UltraViolet Direction And Ranging}
  \acro{UT}[UT]{Unscented Transform}
  \acro{GNSS}[GNSS]{Global Navigation Satellite System}
  \acro{RTK}[RTK]{Real-time kinematic}
  \acro{MOCAP}[mo-cap]{Motion capture}
  \acro{ROS}[ROS]{Robot Operating System}
  \acro{MPC}[MPC]{Model Predictive Control}
  \acro{MBZIRC}[MBZIRC 2020]{Mohamed Bin Zayed International Robotics Challenge 2020}
  \acro{MBZIRC19}[MBZIRC 2019]{Mohamed Bin Zayed International Robotics Challenge 2019}
  \acro{FOV}[FOV]{Field Of View}
  \acrodefplural{FOV}[FOVs]{Fields of View}
  \acro{ICP}[ICP]{Iterative closest point}
  \acro{FSM}[FSM]{Finite-State Machine}
  \acro{IMU}[IMU]{Inertial Measurement Unit}
  \acro{EKF}[EKF]{Extended Kalman Filter}
  \acro{LKF}[LKF]{Linear Kalman Filter}
  \acro{POMDP}[POMDP]{Partially Observable Markov Decision Process}
  \acro{KF}[KF]{Kalman Filter}
  \acro{COTS}[COTS]{Commercially Available Off-the-Shelf}
  \acro{ESC}[ESC]{Electronic Speed Controller}
  \acro{lidar}[LiDAR]{Light Detection and Ranging}
  \acro{SLAM}[SLAM]{Simultaneous Localization and Mapping}
  \acro{SEF}[SEF]{Successive Edge Following}
  \acro{IEPF}[IEPF]{Iterative End-Point Fit}
  \acro{USAR}[USAR]{Urban Search and Rescue}
  \acro{SAR}[SAR]{Search and Rescue}
  \acro{ROI}[ROI]{Region of Interest}
  \acro{WEC}[WEC]{Window Edge Candidate}
  \acro{UAS}[UAS]{Unmanned Aerial System}
  \acro{VIO}[VIO]{Visual-Inertial Odometry}
  \acro{DOF}[DOF]{Degree of Freedom}
  \acrodefplural{DOF}[DOFs]{Degrees of Freedom}
  \acro{LTI}[LTI]{Linear Time-Invariant}
  \acro{FCU}[FCU]{Flight Control Unit}
  \acro{UWB}[UWB]{Ultra-wideband}
  \acro{ICP}[ICP]{Iterative Closest Point}
  \acro{NIS}[NIS]{Normalized Innovations Squared}
  \acro{LRF}[LRF]{Laser Rangefinder}
  \acro{RMSE}[RMSE]{Root Mean Squared Error}
  \acro{VINS}[VINS]{Vision-aided Inertial Navigation Systems}
  \acro{VSLAM}[VSLAM]{Visual Simultaneous Localization and Mapping}
  \acro{NLS}[NLS]{Non-linear Least Squares}
  \acro{NTP}[NTP]{Network Time Protocol}
  \acro{ATE}[ATE]{Absolute Trajectory Error}
  \acro{PUAV}[pUAV]{primary UAV}
  \acro{SUAV}[sUAV]{secondary UAV}
  \acro{NTP}[NTP]{Network Time Protocol}
  \acro{LOS}[LOS]{line-of-sight}
  \acro{MAE}[MAE]{Mean Absolute Error}
\end{acronym}



\bibliographystyle{IEEEtran}
\bibliography{main}


\end{document}